\documentclass[11pt]{article}

\usepackage[margin=1in]{geometry}
\usepackage[T1]{fontenc}
\usepackage[utf8]{inputenc}
\usepackage{lmodern}
\usepackage{booktabs}
\usepackage{amsmath}
\usepackage{amssymb}
\usepackage{graphicx}
\usepackage{multirow}
\usepackage{tabularx}
\usepackage{float}
\usepackage{listings}
\usepackage[colorlinks=true,linkcolor=black,citecolor=black,urlcolor=blue]{hyperref}
\usepackage{algorithm}
\usepackage{algpseudocode}
\usepackage{appendix}
\usepackage{longtable}
\usepackage{authblk}

\title{Cognitive Twins: Investigating Personalized Thinking Model Building and Its Performance Enhancement with Human-in-the-Loop}

\author[1,2]{Wu-Yuin Hwang}
\author[2,*]{Nur Alif Ilyasa}
\author[1,3]{Muhammad Irfan Luthfi}
\author[3]{Yuniar Indrihapsari}

\affil[1]{National Central University, Taoyuan, Taiwan}
\affil[2]{National Dong Hwa University, Hualien, Taiwan}
\affil[3]{Universitas Negeri Yogyakarta, Yogyakarta, Indonesia}
\affil[*]{Corresponding author: \href{mailto:nuralif.ilyasa@gmail.com}{nuralif.ilyasa@gmail.com}}

\date{}

\begin{document}

\maketitle

\begin{abstract}
This paper presents the Personalized Thinking Model (PTM), a hierarchical and interpretable learner representation designed for AI supported education. PTM organizes evidence from learner journals into a five-layer structure covering behavioral instances, behavioral patterns, cognitive routines, metacognitive tendencies, and self-system values. PTM is grounded in Marzano’s New Taxonomy of Educational Objectives and tries to clone learner’s thinking model and build cognitive twin. It was constructed using a pipeline that combines large language model inference (Gemini 2.5 Pro), sentence embeddings, dimensionality reduction, and consensus clustering. This paper evaluates PTM fidelity through three methods applied to 40 participants in a seven-week study. First, automatic evaluation using atomic information point matching yielded an overall F1 score of 74.57\% before human-in-the-loop (HITL) refinement and 75.48\% after refinement. Second, user evaluation using a Likert scale produced mean ratings of 4.26 and 4.30 on a five-point scale for pre and post-HITL conditions respectively. Third, semantic alignment verification showed that topic coherence increased from 0.436 at the behavioral layer to 0.626 at the core value layer, while lexical overlap with journal vocabulary decreased from 0.114 to 0.007 across those same layers. These results suggest that the PTM produces outputs with acceptable fidelity, was generally perceived by users as reflecting their thinking, and showed a pattern consistent with semantic abstraction across layers.
\end{abstract}

\section{Introduction}
\label{sec:introduction}

Effective educational personalization depends not only on what a learner studies, but also on how the learner interprets tasks, regulates effort, and applies knowledge \cite{winne2001}. Systems that store only basic profiles or activity logs provide limited insight into why learners may respond differently to similar prompts \cite{gasevic2016}. To support more individualized assistance, learner models may need to represent not only performance and activity traces but also aspects of reasoning \cite{azevedo2013,roll2011}. This requirement aligns with Marzano's New Taxonomy, which describes learning as involving cognitive processing, metacognitive regulation, and self system motivation \cite{marzano2007,yang2023marzano}. Systems that track only correctness or content exposure provide only partial evidence about how learners plan, monitor, or regulate their work \cite{shute2011,winne2010}.

Existing student modeling research continues to describe cognitive structure assessment as a persistent challenge and suggests that richer cognitive structures can improve representation and interpretability beyond performance-oriented estimation alone \cite{gu2025csg}. Gu et al. \cite{gu2025csg} proposed the Cognitive Structure Generation framework to address a limitation in existing knowledge tracing models, which usually estimate concept-level mastery but do not explicitly represent how learners organize relations among concepts. Takii et al. \cite{takii2024oklm} proposed the Open Knowledge and Learner Model, which integrates behavioral logs with estimates of internal knowledge states. Likewise, Wang et al. \cite{wang2023dynamic} introduced Dynamic Cognitive Diagnosis to improve interpretability in educational deep learning models, arguing that predictive performance alone does not adequately explain the learner's underlying knowledge structure.

Despite these advances, many personalization systems are still described mainly through surface-level metrics rather than through inspectable representations of learner processes \cite{luckin2016}. Luckin et al. \cite{luckin2016} describes a related black-box problem in which an AI system may detect task failure without clearly distinguishing among possible causes such as a knowledge deficit, misconception, or metacognitive difficulty. This limitation supports the need for interpretable approaches that connect observed activity to a more explicit account of learner processes. In this research, the Personalized Thinking Model (PTM) is developed to address these limitations. PTM is positioned as an assistant partner rather than a teaching or coaching agent. It is designed to build an interpretable profile of the learner's thinking and to use that profile to provide tailored scaffolding. By acting as a continuous learning companion, PTM explores the early potentials of Cognitive Twins in education.

The design of the PTM is based on the concept of the digital cognitive twin, which models a learner's internal thinking, planning, and motivation rather than simply copying physical objects. By operating as a persistent digital counterpart, the cognitive twin captures the learner’s foundational knowledge state and continuously adapts as new information is processed. Because the system can maintain and extend this knowledge representation continuously, it is uniquely positioned to assist, guide, and challenge learners to expand their cognitive capabilities. As the learner advances, the cognitive twin dynamically updates its model, establishing a sustainable loop of mutual development where both the learner and the cognitive twin consistently elevate their shared cognition. This unending cycle of mutual improvement extends personalized scaffolding into a sustained practice, paving the way for eternal learning; this is because PTM can continue learning through interacting with the surroundings even though the owner of PTM dies. Eternal learning is defined as a systematic approach to capturing and perpetuating human wisdom through digital repositories, focusing on research into data processing and model generation that transcends the natural limitations of human knowledge sharing \cite{luthfi2024elmts}. The system design is linked to the evaluation strategy. PTM organizes evidence from learner journals into five structured layers, grounded conceptually in Marzano's New Taxonomy \cite{marzano2007}. This study evaluates the fidelity and validity of the PTM as an interpretable personalized model. It examines whether the PTM can represent learner thinking patterns with acceptable fidelity, whether users perceive the generated profiles as reflecting their thinking, and whether the multilayer structure shows a consistent pattern of semantic abstraction and structural integrity from behavioral evidence to higher-level interpretation.

\section{Related Work}
\label{sec:related-work}

\subsection{Theoretical Foundations}

The PTM architecture is grounded primarily in Marzano's New Taxonomy \cite{marzano2007,yang2023marzano}, which defines three interacting systems. The Cognitive System governs task execution, such as vocabulary retrieval and short-term planning. The Metacognitive System concerns goal setting and regulatory monitoring over time. The Self-System encompasses the belief networks and motivational priorities related to task engagement. In this research, this taxonomic structure provides a blueprint for allocating specific evidence tokens across different layers of abstraction. While Bloom's Taxonomy \cite{bloom1956,anderson2001} serves as a supporting parallel indicating that simpler tasks scaffold complex ones, Marzano's tripartite division of self, metacognition, and cognition offers better coverage for modeling a learner's broader characteristics.

This design is also based on the concept of the digital 'cognitive' twin, which models a learner's internal thinking, planning, and motivation rather than simply copying physical objects. Within the context of this research, Digital Twin theory is applied to the cognitive domain to create a digital 'cognitive' twin. Rather than representing a physical object, this twin serves as a structural clone of the user's thinking patterns, cognitive utilization, and metacognitive strategies. By integrating multi-layered evidence into a synchronized representation, the PTM functions as a dynamic digital replica that tracks and simulates the user's unique cognitive architecture.

\subsection{Interpretable Learner Modeling}

Educational AI often navigates a tradeoff between predictive performance and transparent interpretability. When the internal representation cannot be inspected, learners cannot confirm or correct the assumptions the system is making about them \cite{luckin2016,zawacki2019}. Open learner model research fundamentally maintains that visible diagnostic representations foster reflection, trust, and negotiation by users \cite{bullkay2010,bullkay2007}. While systems such as the Open Knowledge and Learner Model visually map concept mastery \cite{takii2024oklm}, and Dynamic Cognitive Diagnosis adds interpretability to knowledge tracing \cite{wang2023dynamic}, these frameworks typically limit inspection to curriculum topics. By providing an inspectable representation stretching across abstract cognitive dimensions, the PTM attempts to operationalize transparency more deeply.

\subsection{Factual Accuracy and Output Grounding}

Evaluating generated assertions in AI platforms often requires decomposing paragraphs into verifiable units. Min et al.\ introduced FActScore, an approach separating generated claims into atomic facts and verifying them against trusted sources \cite{min2023factscore}. Mathematically, the FActScore for a single generated response ($y$) is defined as the percentage of its generated atomic facts ($\mathcal{A}_y$) that are supported by a designated knowledge source ($\mathcal{C}$):
\begin{equation}
f(y) = \frac{1}{|\mathcal{A}_y|} \sum_{a \in \mathcal{A}_y} [a \text{ is supported by } \mathcal{C}] = \text{Precision}
\end{equation}

A similar logic is applied in the RAGAS framework, which evaluates the faithfulness of retrieval-augmented generation systems by decomposing answers into individual statements and verifying their entailment against retrieved context \cite{es2023ragas}. Mathematically, the faithfulness score ($F$) is calculated as the ratio of generated statements ($S$) that are supported by the context ($C$):
\begin{equation}
F = \frac{|S_{\text{supported by } C}|}{|S_{\text{total}}|} = \text{Precision}
\end{equation}

While these frameworks focus exclusively on precision and are typically demonstrated on open-domain biographical datasets from Wikipedia, this research evaluates participant-specific private journal content. To account for both minimizing hallucination (precision) and avoiding omission (recall) in this personalized context, this research adopts an atomic verification mechanism summarized by the F1 metric, adopting the 0.70 benchmark often cited in complex information extraction evaluations \cite{chinchor1998}.

\section{Personalized Thinking Model (PTM) Architecture}
\label{sec:system-design}

The Personalized Thinking Model (PTM) is the main AI component in this research. It organizes journal and interaction evidence into a layered representation of each user's behavioral pattern, cognitive utilization, metacognitive, and core value. Figure~\ref{fig:ptm-overview} in Appendix~\ref{app:ptm-arch} provides an overview of the PTM architecture.

The PTM uses a five-layer hierarchical architecture (L0 to L4). Each layer represents a different level of abstraction, from concrete event records to higher-level patterns and values. This organization design is grounded primarily in Marzano's taxonomy, which provides the main theoretical structure for the progression from concrete events to abstract values. Bloom's revised taxonomy is used only as a limited supporting parallel for the lower layers, especially L0 and L1.

The architecture distinguishes between three functional categories. First, the Data Substrate (L0) contains atomic 5W1H event instances extracted directly from raw user narratives. This layer serves as the evidence base from which the L1 Behavioral Pattern is synthesized and to which higher-layer inferences remain traceable. Second, the Foundational Layer (L1) contains the Behavioral Pattern synthesized from clustered L0 instances without analytical-dimension guidance. Finally, the Dimension-based Layers (L2, L3, L4) contain higher-level nodes synthesized through Analytical Dimensions (AD) with layer-specific focus keywords. The analytical dimensions are defined based on the foundational layer data (L1). 

The selection of a five layer architecture was informed by both cognitive theory and computational modeling research. Prior studies on hierarchical cognitive architectures have demonstrated that moderate depth hierarchies, typically comprising four to six layers, achieve an optimal balance between representational expressivity and computational tractability \cite{shapiro2023ace}. Deeper hierarchies enable finer grained separation of cognitive roles, where each layer can specialize in distinct abstraction levels, thereby improving modularity, memory persistence, and abstraction mapping \cite{xu2023neuromorphic}. However, it is also indicates that additional layers beyond this moderate range yield diminishing returns, as increased complexity may harm scalability or cause overfitting without corresponding architectural adaptations \cite{poctik2025semantic}.

Table~\ref{tab:ptm-layers} formally defines these structural tiers. The hierarchical arrangement of these five layers is grounded in established cognitive theory from educational psychology. Marzano's taxonomy provides the primary architectural foundation, while Bloom's revised taxonomy offers a limited lower layer parallel rather than a full one to one mapping across the whole model. The progression from atomic events (L0) through behavioral patterns (L1), routines (L2), reasoned goals (L3), and the core value layer (L4) follows a movement from concrete evidence to abstract interpretation.

The Behavioral Instance (L0) layer captures specific, timestamped events extracted directly from user journal entries using the 5W1H framework of what, when, where, who, why, and how. These L0 nodes form the evidential foundation for all higher order inferences and map to Marzano's Knowledge Domain. The Behavioral Pattern (L1) layer synthesizes L0 instances into recurring behavioral patterns through semantic clustering. L1 nodes abstract away specific dates to capture generalizable observations, mapping to Marzano's Cognitive System for Comprehension.

The Cognitive Utilization (L2) layer applies analytical dimensions to L1 patterns to represent how they function as strategic routines. L2 nodes capture habit formation, trigger response mechanisms, and routine maintenance strategies, mapping to Marzano's Cognitive System for Knowledge Utilization. The Metacognitive (L3) layer represents the goal prioritization, planning logic, and decision rationale used to interpret L2 strategies. L3 nodes capture the interpretive why behind the how, mapping to Marzano's Metacognitive System. Finally, the Core-Value (L4) layer synthesizes L3 reasoning into fundamental beliefs and deep motivations. These represent the most stable and abstract aspects of user cognition, mapping to Marzano's Self-System.

Marzano's taxonomy explicitly addresses metacognition and the self-system, which are dimensions absent from Bloom's original framework. The alignment is as follows. L0 maps to the Knowledge Domain; L1 maps to the Cognitive System for Comprehension; L2 maps to the Cognitive System for Knowledge Utilization; L3 maps to the Metacognitive System; and L4 maps to the Self System.

\begin{table}[htbp]
\centering
\caption{PTM Knowledge Graph Layer Definitions}
\label{tab:ptm-layers}
\resizebox{\textwidth}{!}{\begin{tabular}{@{}clp{5cm}ll@{}}
\toprule
\textbf{Layer} & \textbf{Name} & \textbf{Focus} & \textbf{Bloom's Parallel} & \textbf{Marzano's Parallel} \\
\midrule
L0 & Behavioral Instance & \raggedright Atomic 5W1H events (what, when, where, who, why, how) & Remember & Knowledge Domain \\
L1 & Behavioral Pattern & \raggedright Recurring observable patterns synthesized from L0 clusters & Understand & Cognitive System (Comprehension) \\
L2 & Cognitive Utilization & \raggedright Routine maintenance, habit formation, trigger response & Apply & Cognitive System (Utilization) \\
L3 & Metacognitive & \raggedright Goal prioritization, planning logic, decision rationale & -- & Metacognitive System \\
L4 & Core Value & \raggedright Deep motivations, fundamental beliefs, problem solving philosophy & -- & Self System \\
\bottomrule
\end{tabular}}
\end{table}

Examples of complete node structures at each layer, including Behavioral Pattern (L1), Cognitive Utilization (L2), Metacognitive (L3), and Core Value (L4), are provided in Appendix \ref{app:ptm-data-sample}. These examples illustrate how the model functions by cloning the human thinking model for learning and beyond, providing a structured replica of the user's cognitive processes.

\begin{figure}[H]
    \centering
    \includegraphics[width=0.55\linewidth]{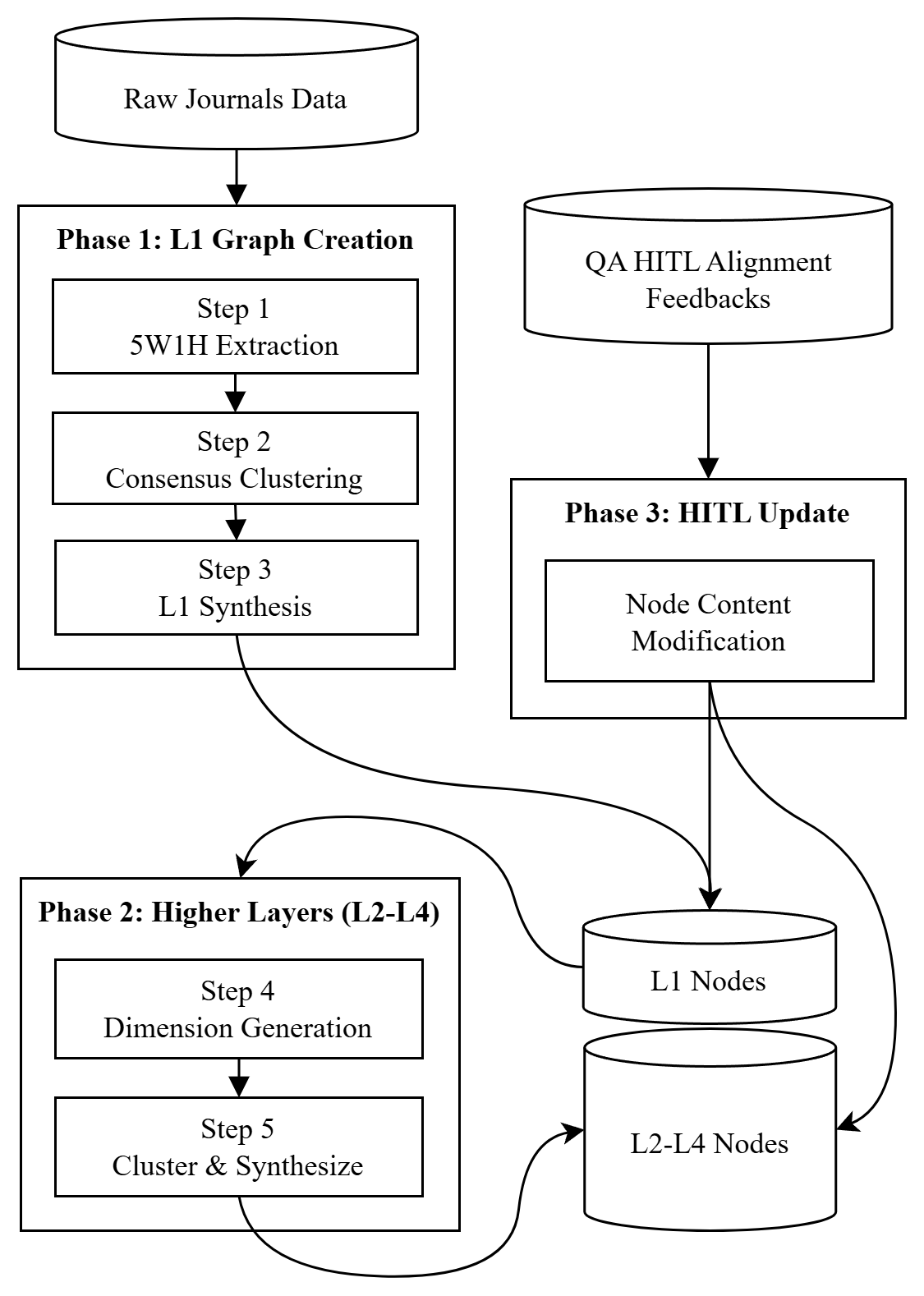}
    \caption{PTM Model Construction Pipeline Overview}
    \label{fig:pipeline-flow}
\end{figure}

The model is generated iteratively using a three phase pipeline divided into five distinct process stages. Figure~\ref{fig:pipeline-flow} provides an overview of the construction pipeline.

\subsection{Phase 1: Data Ingestion and L1 Graph Creation}

\begin{figure}[H]
    \centering
    \includegraphics[width=0.85\linewidth]{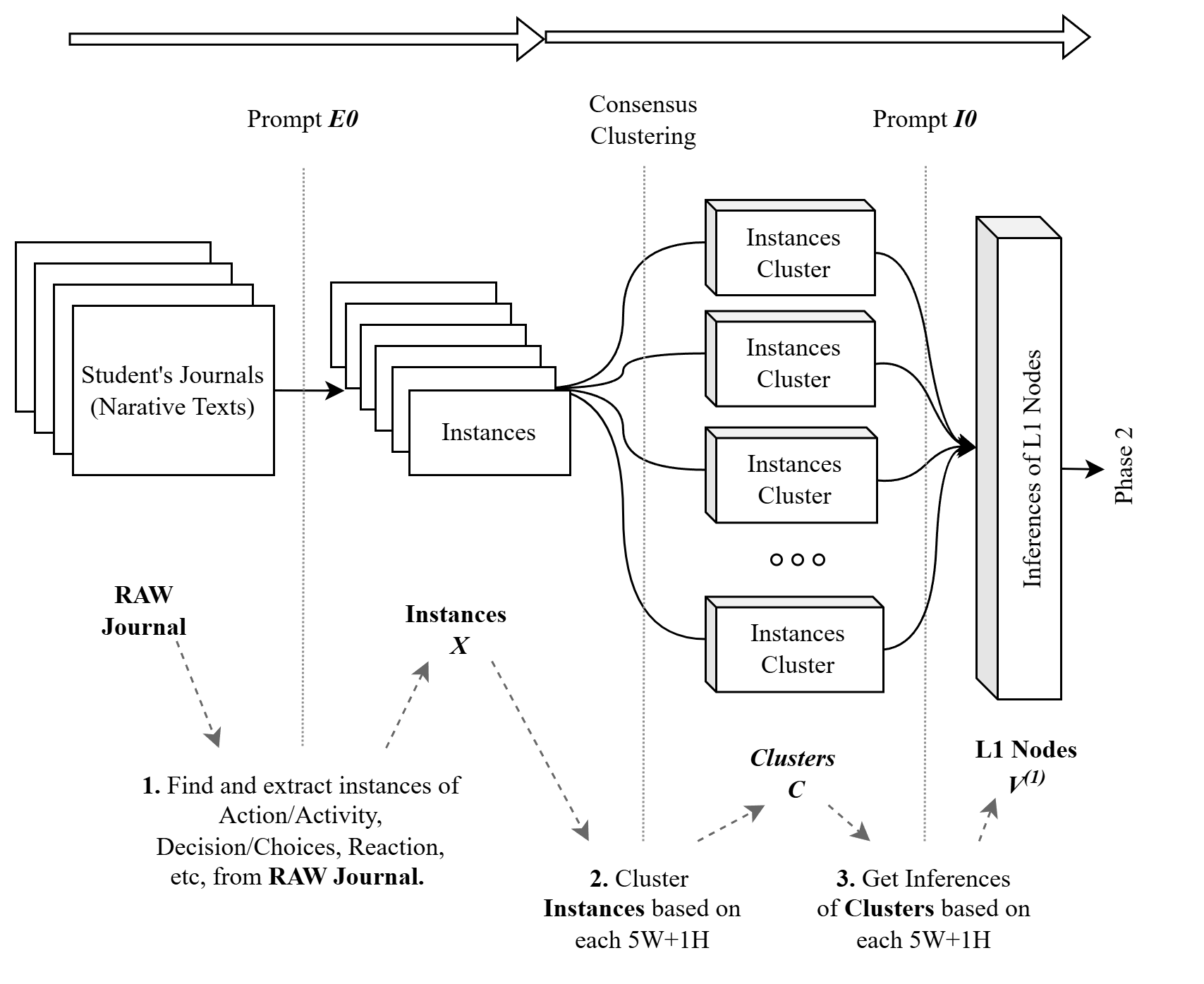}
    \caption{Phase 1: Raw Data to L1 Behavioral Node}
    \label{fig:raw-l1}
\end{figure}

The first phase processes raw user data to create the foundational layers of the knowledge data: L0 (Behavioral Instance) and L1 (Behavioral Pattern). This phase covers data extraction from journal entries, semantic vectorization, consensus clustering, and node synthesis, as illustrated in Figure~\ref{fig:raw-l1}.

Stage 1 focuses on data ingestion and structured extraction to form the L0 base. The pipeline ingests unstructured journal entries and invokes the L0 Extraction prompt (E0) to extract discrete behavioral facts using the 5W1H framework (What, When, Where, Who, Why, and How). Using Google's Gemini 2.5 Pro, the narrative text is parsed into atomic, timestamped JSON instances that preserve the original context while providing a structured foundation for pattern discovery.

Stage 2 performs foundational synthesis via weighted consensus clustering to identify recurring L1 Behavioral Patterns. Let $X$ represent the set of L0 instances. To identify robust patterns, the pipeline runs independent base clusterings for each 5W1H attribute. Each attribute value is converted into dense vector representations using Sentence BERT (all MiniLM L6 v2) \cite{reimers2019}. Dimensionality reduction is applied using UMAP to reduce the data to 25 dimensions, resolving distance concentration issues in high-dimensional spaces \cite{Amblard2022, feldbauer2019comprehensive}. Reduced embeddings are clustered using HDBSCAN to identify groups of varying densities \cite{campello2013}. 

After the six base clusterings are completed, the pipeline constructs a pairwise consensus matrix. For each pair of instances, the matrix calculates a weighted co-occurrence score based on how many base clusterings placed the two instances in the same cluster. An attribute weight function $w(a)$ is defined to reflect the primary importance of the core activity (What = 2, others = 1). The raw weighted co-occurrence score $R(x_i, x_j)$ between any two instances is calculated as the sum of the weights of the attributes for which they share the same base cluster:
\begin{equation}
    R(x_i, x_j) = \sum_{a \in A} w(a) \cdot \mathcal{I}(\pi_a(x_i) = \pi_a(x_j))
\end{equation}
To distinguish genuine inter-day patterns from isolated, intensive single-day sessions, a temporal penalty function $P(x_i, x_j)$ is introduced, which subtracts 2 points if the instances occur on the same date. The final similarity score $S(x_i, x_j)$ populating the consensus matrix is:
\begin{equation}
    S(x_i, x_j) = R(x_i, x_j) - P(x_i, x_j)
\end{equation}
Pairs meeting the threshold $\tau=4$ form stable clusters. This threshold is calibrated to ensure that only stable, inter day patterns are recognized as genuine behavioral characteristics \cite{nguyen2007}. Let $C$ be the set of these stable clusters, defined as:
\begin{equation}
    C = \{c_1, c_2, \dots, c_M\}, \quad \text{where} \quad c_m \subseteq X
\end{equation}
Each cluster $c_m$ is then passed to the L1 Node Synthesis prompt (IO), which acts as a synthesis function $f_{\text{IO}}(c_m)$. This function instructs the LLM to analyze the clustered L0 instances and synthesize them into distinct L1 Behavioral Pattern nodes. Let $V^{(1)}$ be the set of synthesized L1 nodes, defined as:
\begin{equation}
    V^{(1)} = \{v_1^{(1)}, v_2^{(1)}, \dots, v_M^{(1)}\}, \quad \text{where} \quad v_m^{(1)} = f_{\text{IO}}(c_m)
\end{equation}
Figure~\ref{fig:raw-l1} illustrates this Phase 1 process.

\begin{algorithm}[H]
\caption{Applied Weighted Consensus Clustering with Temporal Penalty}\label{alg:consensus}
\begin{algorithmic}[1]
\Require Set of L0 Instances $X$, threshold $\tau = 4$
\Ensure Set of L1 Behavioral Pattern clusters $C$
\State \textbf{Phase 1: Base Clustering}
\For{each 5W1H attribute $a$}
    \State Generate SBERT embeddings and reduce to 25 dimensions via UMAP
    \State Apply HDBSCAN to generate assignments $\pi_a$
\EndFor
\State \textbf{Phase 2: Consensus Matrix}
\For{each pair $(x_i, x_j) \in X \times X$}
    \State Calculate raw score $R_{ij}$ using weights $w(a)$ (What=2, others=1)
    \State Apply penalty $P_{ij}=2$ if dates match, else $P_{ij}=0$
    \State $S_{ij} \gets R_{ij} - P_{ij}$
\EndFor
\State \textbf{Phase 3: Final Synthesis}
\State Group instances where $S_{ij} \ge \tau$ into clusters $C$
\end{algorithmic}
\end{algorithm}

\subsection{Phase 2 of Higher Layer Synthesis (L2, L3, L4)}

The second phase constructs higher-order layers through progressive abstraction. Unlike Phase 1, which uses consensus clustering to discover patterns from raw behavioral data, Phase 2 uses analytical dimensions to guide the synthesis of L1 Behavioral Patterns into L2 (Cognitive Utilization), L3 (Metacognitive), and L4 (Core Value) nodes. Let $L_n$ denote the target higher layer, where $n \in \{2, 3, 4\}$. The dimension-based synthesis process transforms the set of L1 nodes $V^{(1)}$ into a new set of higher-layer nodes $V^{(n)}$.

The synthesis of L1 Behavioral Patterns into higher-layer nodes begins with generating the analytical dimensions using the Dimension Generation prompt (GD). Rather than analyzing the entire set of L1 Behavioral Patterns $V^{(1)}$, the prompt acts as a function $f_{\text{GD}}(\hat{V}^{(1)})$ that analyzes a uniformly sampled subset. Let $\hat{V}^{(1)} \subset V^{(1)}$ represent this randomly sampled subset containing exactly 50 nodes. This specific sample size of 50 nodes is tuned to balance generalizability with the context-window limitations of the LLM. Figure~\ref{fig:dimension-gen} illustrates the dimension generation process. Although Gemini 2.5 Pro possesses a one-million-token context window, processing an excessively large number of tokens complicates the reasoning task and can degrade the accuracy of information extraction \cite{liu2024lost}. The subset of 50 nodes provides a sufficiently representative base to generalize the whole set without exceeding optimal context thresholds. Let $D^{(n)}$ be the set of analytical dimensions specific to layer $L_n$:

\begin{figure}[H]
    \centering
    \includegraphics[width=0.7\linewidth]{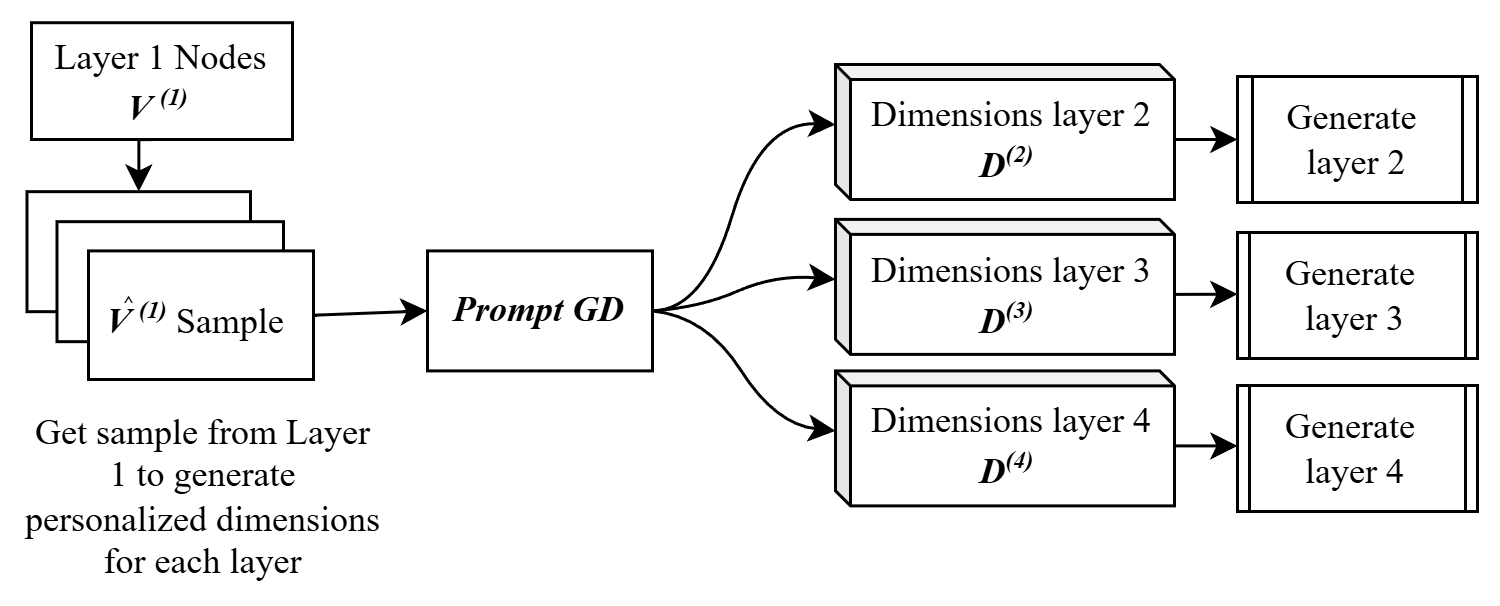}
    \caption{Phase 2: Dimension Model Generation Process}
    \label{fig:dimension-gen}
\end{figure}
\begin{equation}
    D^{(n)} = \{d_1^{(n)}, d_2^{(n)}, \dots, d_K^{(n)}\}
\end{equation}
The complete set of generated dimensions $D$ across all three higher layers is formally defined as:
\begin{equation}
    D = f_{\text{GD}}(\hat{V}^{(1)}) = D^{(2)} \cup D^{(3)} \cup D^{(4)}
\end{equation}
The prompt enforces the hierarchical structure by specifying layer specific keywords and focus areas. The L2 Dimensions $D^{(2)}$ focus on habits, routines, and behavioral triggers. The L3 Dimensions $D^{(3)}$ focus on reasoning, goals, priorities, and planning. Finally, the L4 Dimensions $D^{(4)}$ focus on Core Value, motivations, and problem solving philosophy. 

Following the dimension generation, the nodes are grouped into semantically coherent clusters based on each specific analytical dimension $d \in D^{(n)}$. Unlike the initial baseline clustering which utilizes density-based algorithms on text embeddings, the subsequent grouping for higher layers relies exclusively on prompt-based clustering. To optimize the language model processing and reduce token complexity, the Dimensional Clustering prompt (CD) takes only the descriptive labels of the previous-layer nodes as input, rather than processing the full node contents. 

Let $f_{\text{CD}}$ denote this prompt-based clustering function applied to the previous-layer nodes for a given dimension $d$. The prompt evaluates the input node labels and groups them into semantically coherent clusters based on the specific analytical dimension. Let $C_d^{(n)}$ be the set of dimension-specific clusters formed under dimension $d$, formally defined as:
\begin{equation}
    C_d^{(n)} = \{c_{d,1}^{(n)}, c_{d,2}^{(n)}, \dots, c_{d,P}^{(n)}\}
\end{equation}
Each cluster contains a subset of the input nodes, where $p$ serves as the integer index identifying the specific cluster within that dimension. Each cluster also receives a descriptive label that characterizes the shared commonality of its grouped nodes.

\begin{figure}[H]
    \centering
    \includegraphics[width=0.85\linewidth]{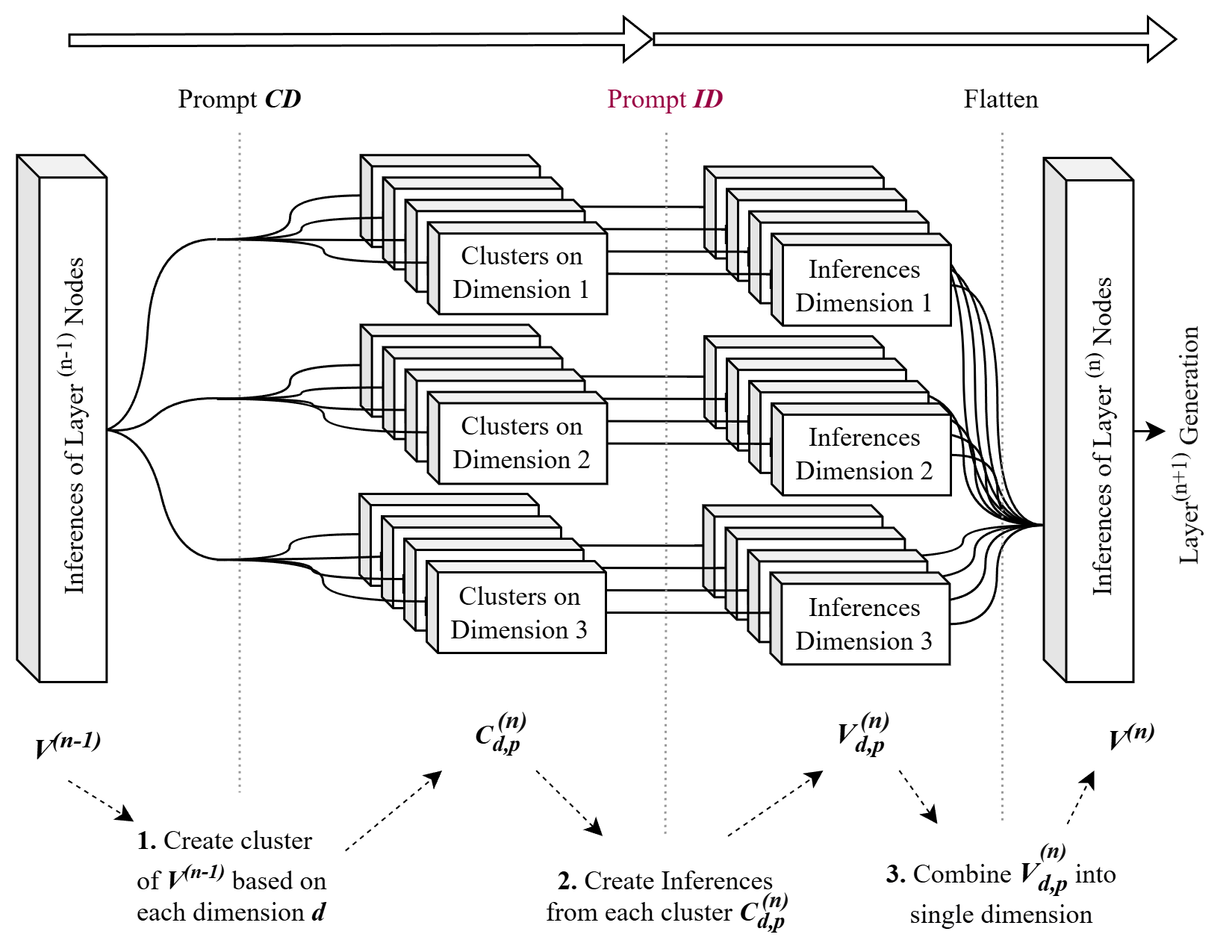}
    \caption{Phase 2: L2-L4 Node Generation Flow}
    \label{fig:l2n-gen}
\end{figure}

The Graph Based Synthesis prompt (ID) generates abstract insights from the clustered patterns. Figure~\ref{fig:l2n-gen} illustrates this recursive synthesis detail. This prompt acts as a generative function $f_{\text{ID}}(c_{d,p}^{(n)}, d)$. Given a specific analytical dimension $d$ and its corresponding cluster $c_{d,p}^{(n)}$ (where $p$ represents the specific cluster index), the LLM synthesizes one to three higher level interpretations that explain why these patterns appear together. Let $V^{(n)}_{d,p}$ be the set of generated higher layer nodes for that cluster, defined as:
\begin{equation}
    V^{(n)}_{d,p} = f_{\text{ID}}(c_{d,p}^{(n)}, d)
\end{equation}
The total set of nodes for layer $L_n$, denoted as $V^{(n)}$, is the union of all nodes generated across all dimensions in $D^{(n)}$ and their respective clusters. This is formally expressed as:
\begin{equation}
    V^{(n)} = \bigcup_{d \in D^{(n)}} \bigcup_{p} V^{(n)}_{d,p}
\end{equation}
This helps higher layer nodes state generalizable interpretive patterns rather than isolated actions.

\subsection{Phase 3 of Model Refinement (HITL)}
\label{subsubsec:phase3-hitl-method}
The third phase incorporates human feedback to refine the model. This human in the loop (HITL) stage addresses a practical limitation of automated modeling from journal evidence alone. Higher layer nodes, especially L3 and L4, require interpretation beyond directly observable behavior, so they can be harder to verify from text alone. Following the general HITL rationale discussed by Amershi et al. \cite{amershi2014} and Holzinger \cite{holzinger2016}, user feedback is treated in this research as a review signal for checking whether the generated nodes are acceptable, understandable, and in need of revision.

The HITL process employed in this study is presented to the users with fact checking questions aligned with the principle of accessible model inspection that does not require technical expertise. Let $v \in V^{(n)}$ represent any existing node in the generated PTM graph. The system generates a fact checking question $q$ based on the information contained in node $v$. This interface allows users to review model statements without needing to interpret the underlying graph structure. User feedback $f$ is collected through their response to the question $q$. When users provide corrections or additional context through feedback $f$, the Node Refinement prompt (NR) guides the LLM in updating the existing node. This refinement process acts as a function $f_{\text{NR}}(v, f)$ that outputs an updated node $v'$. Thus, the refined node is formally expressed as:
\begin{equation}
    v' = f_{\text{NR}}(v, f)
\end{equation}
The prompt asks the model to integrate prior and new content into one clear description, retain necessary source meaning, and return only the revised JSON content. Samples of the HITL interaction data, including the types of questions generated and the varying levels of participant response complexity, are provided in Appendix \ref{app:hitl-qa}.

\begin{table}[htbp]
\centering
\caption{PTM Pipeline Hyperparameters}
\label{tab:hyperparameters}
\small
\begin{tabular}{@{}lllp{5.5cm}@{}}
\toprule
\textbf{Component} & \textbf{Parameter} & \textbf{Value} & \textbf{Rationale} \\
\midrule
LLM & Model & Gemini 2.5 Pro & Selected for reasoning capability \cite{gemini2025report} \\
LLM & Temperature & 0.0 & Deterministic, reproducible responses \\
Embedding & Model & all MiniLM L6 v2 & Semantic vectorization \cite{reimers2019} \\
Clustering & Algorithm & HDBSCAN + Consensus & Multi attribute stability \\
UMAP & Components & 25 & Curse of dimensionality reduction \\
Consensus & Threshold ($\tau$) & 4 & Inter day pattern stability \\
\bottomrule
\end{tabular}
\end{table}

Table~\ref{tab:hyperparameters} summarizes the configuration constants used throughout the pipeline. The LLM temperature is set to 0.0 to ensure deterministic, reproducible outputs across all generation calls. The embedding model was selected for its balance of inference speed and semantic quality. Clustering parameters were tuned to require minimal recurring evidence before forming a behavioral pattern. UMAP components were set to 25 to resolve the curse of dimensionality and improve density contrast for HDBSCAN \cite{Amblard2022, feldbauer2019comprehensive}. The minimum consensus score $\tau$ was set to 4 to ensure that only stable inter day patterns are synthesized.

\section{Methodology}
\label{sec:methodology}

Evaluation data was collected during a seven week field study at Universitas Negeri Yogyakarta, Indonesia, examining undergraduates in an Information Technology degree program. For the core evaluation of the PTM itself, the analysis utilized the 40 participants. These students experienced complete graph construction and HITL refinement. All were English as a Foreign Language learners aged 18 to 24. A baseline phase (Weeks 1 to 5) aggregated initial foundational data, followed by an intervention phase (Weeks 6 to 7) applying PTM scaffolding in a live journal writing platform.

\subsection{Automatic Evaluation}

The automatic evaluation assessed whether the PTM could accurately answer factual questions about user records, drawing from atomic evaluation methodology \cite{min2023factscore, es2023ragas}. For each participant, a layer balanced question set averaging 27.5 items was extracted from random multi day windows along with known ground truth journal sentences. The system was restricted to answering purely using its synthesized PTM context. The generated response and the ground truth were then decomposed into semantic atomic clauses. Scoring was calculated by determining true positive counts (matched clauses), false positives (hallucinated claims not in the ground truth), and false negatives (ground truth clauses missed by the prediction). The F1 score was used as the primary metric because the evaluation needed to reflect both correctness and coverage. Using the F1 score provides a more balanced assessment of PTM outputs under Atomic Information Point Matching than precision alone.

\subsection{User Evaluation}
\label{subsec:user-evaluation}

User acceptance was tabulated through an 18 item randomized questionnaire surveying synthesized statements from L1 through L4. Participants used a five point Likert scale, where 1 indicated strong disagreement and 5 indicated strong agreement, to rate their sense of personal identification with the statements. These scores were recorded before the human verification process (pre HITL) and after users had manually adjusted aspects of the graph (post HITL). Qualitative interviews with four participants were also conducted as a follow up step to further interpret student perception and provide evidence for the numerical ratings.

\subsection{Semantic Hierarchy Evaluation}

To determine whether successive layers demonstrated climbing abstraction levels, both internal consistency (internal semantics) and external lexical grounding (external grounding) were calculated.

Internal semantics refers to whether the generated nodes form a cohesive and organized system. This is evaluated through four indicators. First, Topic Coherence ($C_v$) measures within-topic semantic consistency. It evaluates thematic unity and internal consistency by measuring how reliably words co-occur within the dataset \cite{mimno2011optimizing, roder2015exploring}. Second, Similarity evaluates how closely related the themes are within each layer. While text data naturally tends to clump together mathematically (anisotropy), the relative increase in similarity across layers indicates denser semantic grouping. Third, the Silhouette Coefficient evaluates clustering quality by measuring cluster separation. Although traditional benchmarks expect scores above 0.50, the vast variety of unique vocabulary in high-dimensional text data naturally suppresses silhouette magnitudes due to the curse of dimensionality. Therefore, consistently positive silhouette values are used to confirm valid thematic separation. Fourth, Topic Count identifies the number of semantic themes per layer to ensure the model preserves clear distinctions without fragmenting into unstable micro topics.

External grounding examines whether these abstractions remain connected to the original journal content. This is measured using Jaccard Similarity indices between the PTM node vocabulary and source journal vocabulary. In concept mapping research, a Jaccard overlap of at least 0.10 is commonly used as a technical threshold to establish sufficient grounding for automated knowledge maps extracted from unstructured text \cite{takii2024oklm}. The objective was to ascertain that L4 nodes were not merely copying journal words literally, but restructuring them conceptually through abstraction.

\section{Results}
\label{sec:results}

\subsection{PTM Distribution and Properties}

All 40 participants in the Experimental Group had complete and valid models across all four layers from L1 to L4. The structural analysis was conducted on the pre HITL PTM state. This is because HITL refinement updates node content but does not change node counts or linkage structure. Table~\ref{tab:ptm-nodes} presents the node distribution across cognitive layers prior to model refinement.

\begin{table}[H]
\centering
\caption{PTM Node Distribution by Cognitive Layer (Pre HITL, $n = 40$ users)}
\label{tab:ptm-nodes}
\small
\begin{tabular}{@{}lrrrr@{}}
\toprule
\textbf{Layer} & \textbf{Mean} & \textbf{SD} & \textbf{Total} & \textbf{Range} \\
\midrule
L1: Behavioral Pattern & 73.20 & 19.07 & 2,928 & 41--144 \\
L2: Cognitive Utilization & 36.80 & 7.11 & 1,472 & 24--73 \\
L3: Metacognitive & 27.57 & 3.19 & 1,103 & 18--42 \\
L4: Core Value & 18.02 & 1.44 & 721 & 12--21 \\
\midrule
\textbf{Total} & 155.59 & 27.89 & 6,224 & -- \\
\bottomrule
\end{tabular}
\end{table}

The node distribution data reveals a decrease from L1 to L4, with an approximately 4:1 ratio between the behavioral layer (M = 73.20) and the core value layer (M = 18.02). This distribution reflects the theoretical structure of hierarchical cognition where specific daily behaviors are numerous and variable while the Core Value attributes are fewer and more stable \cite{marzano2007}. Furthermore, the linkage density analysis in Table~\ref{tab:ptm-linkage} indicates that nodes in the higher layers are linked to multiple observations from the layers below them. The model achieved an overall mean in-degree of 3.06 across layers L2 to L4. This value suggests that nodes in the higher layers were generally linked to about three lower-layer nodes rather than to isolated observations. This pattern is consistent with a hierarchical structure that does not rely on single observations alone.

\begin{table}[htbp]
\centering
\caption{PTM Linkage Density by Cognitive Layer (Pre HITL, $n = 40$ users)}
\label{tab:ptm-linkage}
\small
\begin{tabular}{@{}lrrrr@{}}
\toprule
\textbf{Layer Pair} & \textbf{Mean In Degree} & \textbf{SD} & \textbf{Total Links} & \textbf{Range} \\
\midrule
L1 to L2 & 3.04 & 0.51 & 4,475 & 1.9--4.1 \\
L2 to L3 & 2.67 & 0.40 & 2,945 & 2.0--3.8 \\
L3 to L4 & 3.47 & 0.44 & 2,502 & 2.2--4.4 \\
\midrule
Overall & 3.06 & 0.45 & 9,922 & 1.9--4.4 \\
\bottomrule
\end{tabular}
\end{table}

Table~\ref{tab:ptm-linkage} reports the inter-layer connectivity of the PTM graph. Mean In-Degree indicates the average number of lower-layer nodes connected to each upper-layer node, and Total Links reports the aggregate edge count across all participants. Graph connectivity evaluations recorded 9,922 total inter-layer links. On average, each upper-layer node relied on 3.06 discrete supporting elements immediately below it. This indicates that abstract representations possessed pluralistic grounding evidence rather than forming from single occurrences.

\begin{table}[htbp]
\centering
\caption{HITL Refinement Input by Marzano Level}
\label{tab:hitl-descriptive}
\small
\begin{tabular}{@{}lrrrr@{}}
\toprule
\textbf{Layer} & \textbf{Count} & \textbf{Total Words} & \textbf{Avg. Words} & \textbf{SD} \\
\midrule
L1: Behavioral Pattern & 260 & 5,436 & 20.91 & 11.75 \\
L2: Cognitive Utilization & 201 & 6,530 & 36.49 & 15.04 \\
L3: Metacognitive & 143 & 5,282 & 32.94 & 14.34 \\
L4: Core Value & 116 & 4,593 & 30.59 & 16.95 \\
\midrule
Total & 720 & 21,841 & 30.23 & 14.15 \\
\bottomrule
\end{tabular}
\end{table}

Table~\ref{tab:hitl-descriptive} reports the volume of textual feedback collected during the HITL phase. Count indicates the total number of node reviews performed across all participants at each layer. Total Words and Avg. Words measure the cumulative and per review verbosity of participant corrections. The HITL interactive review processed 720 discrete node assessments. Participants used longer sentences when modifying Cognitive Utilization strategies (36.49 words on average) compared to correcting Behavioral actions (20.91 words on average). This pattern suggests that conceptual levels required more nuanced text to adequately adjust.

\subsection{Automatic Fidelity Results}

\begin{table}[htbp]
\centering
\caption{Overall Automatic Evaluation Performance: Before vs. After HITL Refinement}
\label{tab:automatic-overall}
\small
\begin{tabular}{@{}lrrrrrrrrr@{}}
\toprule
\textbf{Condition} & \textbf{Prec.} & \textbf{Recall} & \textbf{F1} & \textbf{SD} & \textbf{TP} & \textbf{FP} & \textbf{FN} & \textbf{$t$} & \textbf{$p$} \\
\midrule
Before HITL & 72.50\% & 76.77\% & 74.57\% & 3.38\% & 3,628 & 1,376 & 1,098 & \multirow{2}{*}{--1.53} & \multirow{2}{*}{0.135} \\
After HITL & 73.69\% & 77.35\% & 75.48\% & 3.40\% & 3,638 & 1,299 & 1,065 & & \\
\midrule
Improvement & +1.19 pp & +0.58 pp & +0.91 pp & -- & +10 & --77 & --33 & -- & -- \\
\bottomrule
\end{tabular}
\par\smallskip
\footnotesize\textit{Note: $t$ and $p$ represent the difference of before and after.}
\end{table}

In Table~\ref{tab:automatic-overall}, Precision measures the proportion of predicted atomic points that match ground truth, Recall measures the proportion of ground truth points captured by the prediction, and F1 is the harmonic mean of both. TP, FP, and FN are aggregated counts of true positives, false positives, and false negatives across all evaluation items. Across 4,726 distinct semantic facts, the overall F1 score stood at 74.57\% before HITL and 75.48\% after HITL. Both measurements exceed the 0.70 acceptability threshold. The paired difference ($t = -1.53$, $p = 0.135$) was not statistically significant. The numerical improvement was primarily driven by a reduction in false positives (from 1,376 to 1,299), suggesting that HITL refinement helped reduce hallucinated content rather than recover additional missed information.

\subsection{Layer-by-Layer Results}
\label{subsec:automatic-layer-results}

Table~\ref{tab:automatic-layer-detailed} provides the detailed breakdown of performance metrics for each layer.

\begin{table}[H]
\centering
\caption{Detailed Performance Metrics by PTM Layer}
\label{tab:automatic-layer-detailed}
\small
\begin{tabular}{@{}llrrrrrrrrr@{}}
\toprule
\textbf{Layer} & \textbf{Condition} & \textbf{Prec.} & \textbf{Recall} & \textbf{F1} & \textbf{SD} & \textbf{TP} & \textbf{FP} & \textbf{FN} & \textbf{$t$} & \textbf{$p$} \\
\midrule
\multirow{2}{*}{L1}
 & Before HITL & 72.50\% & 76.94\% & 74.65\% & 7.32\% & 891 & 338 & 267 & \multirow{2}{*}{--1.65} & \multirow{2}{*}{0.107} \\
 & After HITL & 75.13\% & 77.74\% & 76.41\% & 7.13\% & 894 & 296 & 256 & & \\
\midrule
\multirow{2}{*}{L2}
 & Before HITL & 72.78\% & 76.55\% & 74.62\% & 6.22\% & 901 & 337 & 276 & \multirow{2}{*}{--1.62} & \multirow{2}{*}{0.112} \\
 & After HITL & 75.62\% & 78.01\% & 76.80\% & 6.37\% & 940 & 303 & 265 & & \\
\midrule
\multirow{2}{*}{L3}
 & Before HITL & 71.92\% & 76.23\% & 74.01\% & 6.17\% & 927 & 362 & 289 & \multirow{2}{*}{0.18} & \multirow{2}{*}{0.859} \\
 & After HITL & 71.41\% & 76.25\% & 73.75\% & 5.15\% & 899 & 360 & 280 & & \\
\midrule
\multirow{2}{*}{L4}
 & Before HITL & 72.84\% & 77.36\% & 75.03\% & 6.03\% & 909 & 339 & 266 & \multirow{2}{*}{--0.01} & \multirow{2}{*}{0.995} \\
 & After HITL & 72.69\% & 77.42\% & 74.98\% & 6.47\% & 905 & 340 & 264 & & \\
\bottomrule
\end{tabular}
\par\smallskip
\footnotesize\textit{Note: $t$ and $p$ represent the difference of before and after.}
\end{table}

The layer-by-layer analysis summarizes precision, recall, and F1 score across all four PTM layers. As shown in Table~\ref{tab:automatic-layer-detailed}, precision remained above 71 percent and recall above 76 percent for every layer before refinement, while F1 scores ranged from 74.01 to 75.03 percent. After refinement, F1 scores ranged from 73.75 to 76.80 percent. All values remained above the 0.70 threshold \cite{chinchor1998,min2023factscore}.

The Behavioral Pattern and Cognitive Utilization layers showed modest numerical improvements after HITL refinement. The Behavioral Pattern layer reached an F1 score of 76.41 percent (SD = 7.13 percent), and the Cognitive Utilization layer reached 76.80 percent (SD = 6.37 percent). These lower layers contain more concrete content, which may be easier for participants to review during refinement. However, the paired t-tests did not indicate statistically significant change ($L1: t = -1.65, p = 0.107$, $L2: t = -1.62, p = 0.112$).

The Metacognitive and Core Value layers showed little change after refinement. The F1 score for the Metacognitive layer decreased slightly to 73.75 percent (SD = 5.15 percent), while the Core Value layer remained stable at 74.98 percent (SD = 6.47 percent). The paired t-tests likewise did not indicate statistically significant change in these higher layers ($L3: t = 0.18, p = 0.859$, $L4: t = -0.01, p = 0.995$). Given that each user's PTM was refined with only 18 QA items, the available feedback signal may have been limited for these more abstract layers.

One likely reason the third and fourth layers changed little is the abstract nature of the constructs involved. HITL refinement requires users to judge model statements and provide corrective feedback. Metacognitive processes and core values are more abstract than behavioral patterns, which may make them harder for participants to review and revise precisely. Interview data supports this observation. Participant B reported, \textit{``For the higher-level parts, I kept thinking, `Maybe this is partly true,' so I did not know what to change.''} Participant D echoed, \textit{``I did not feel confident editing those statements because they sounded important and abstract. I worried that if I changed them, I might make them less accurate.''} These interview responses suggest that higher-layer descriptions were often perceived as broad rather than obviously false, which limited how precisely participants could revise them. Prior research indicates that abstract concepts are generally harder to communicate verbally than concrete concepts \cite{borghi2017words}. This helps explain why HITL refinement was less responsive at the more abstract PTM layers.

The relatively small change in F1 score is also consistent with the limited amount of HITL refinement data in this study. As defined in Section~\ref{subsec:user-evaluation}, each user's PTM was refined using only 18 feedback items in total. This amount of feedback may have been limited for a task spanning multiple abstraction levels. In addition, the effect of refinement depends not only on the number of feedback items but also on their descriptive quality.

\subsection{User Evaluation Results}

\begin{table}[htbp]
\centering
\caption{User Evaluation: Overall Descriptive Statistics and by PTM Layer}
\label{tab:user-eval}
\small
\begin{tabular}{@{}lcccccccc@{}}
\toprule
\multirow{2}{*}{\textbf{Level}} & \multicolumn{3}{c}{\textbf{Pre HITL}} & \multicolumn{3}{c}{\textbf{Post HITL}} & \multirow{2}{*}{$t$} & \multirow{2}{*}{$p$} \\
\cmidrule(lr){2-4} \cmidrule(lr){5-7}
 & Mean & SD & N & Mean & SD & N & & \\
\midrule
Overall & 4.26 & 0.90 & 720 & 4.30 & 0.88 & 684 & --0.72 & 0.470 \\
\addlinespace
L1: Behavioral & 4.42 & 1.12 & 225 & 4.49 & 1.07 & 212 & --0.65 & 0.517 \\
L2: Cognitive & 4.27 & 0.85 & 248 & 4.30 & 0.83 & 240 & --0.45 & 0.656 \\
L3: Metacognitive & 4.16 & 0.60 & 140 & 4.18 & 0.62 & 132 & --0.33 & 0.739 \\
L4: Core Value & 4.07 & 0.73 & 107 & 4.04 & 0.74 & 100 & 0.25 & 0.804 \\
\bottomrule
\end{tabular}
\end{table}

In Table~\ref{tab:user-eval}, Mean represents the average Likert score on a 1 to 5 scale, where 5 indicates the user perceives the node statement as accurately reflecting their thinking. N indicates the total number of node evaluations at each layer, and the paired t-test compares pre versus post HITL conditions. Mean user agreement remained above the neutral threshold (3.0) across all layers. The highest scores occurred at L1 (4.49 post HITL), confirming that observable activity descriptions translate most directly to recognized behavior. The scores decreased gradually at higher layers, with L4 showing the lowest mean (4.04 post HITL). None of the pre to post HITL differences were statistically significant (all $p > 0.05$).

Participant A highlighted the layer difference in perception: \textit{``Behavioral and cognitive statements were easier because I could connect them to real actions. The self system layer felt more interpretive.''} Participant C noted, \textit{``I accepted some of those statements because they translated my habits into a more general idea. It felt unfamiliar, but not necessarily wrong.''} Another user, Participant B, explained: \textit{``Some phrases sounded more academic than my own language, but I still accepted them because the meaning felt close. I was not judging it by whether it sounded exactly like my diary.''} These interview responses suggest that users prioritized the semantic gist and main themes over literal matching, accepting synthesized insights as long as the core meaning felt accurate.

\subsection{Automatic vs. User-Based Evaluation}
\label{subsec:user-eval-triangulation}

To compare user ratings with automatic evaluation results, this subsection considers the two evaluation streams together. Table~\ref{tab:triangulation} compares these measures. For clarity, the user evaluation percentages in Table~\ref{tab:triangulation} are the normalized forms of the phase means reported in Table~\ref{tab:user-eval}. They do not represent participant percentages.

\begin{table}[H]
\centering
\caption{Comparison of Model Fidelity Measures (Pre HITL vs. Post HITL)}
\label{tab:triangulation}
\small
\begin{tabular}{@{}llrr@{}}
\toprule
\textbf{Metric Source} & \textbf{Measurement Type} & \textbf{Pre HITL} & \textbf{Post HITL} \\
\midrule
Automatic Eval & F1 Score & 74.57\% & 75.48\% \\
User Eval & Likert Scale & 81.56\% & 82.42\% \\
\midrule
Divergence &  & +6.99\% & +6.94\% \\
\bottomrule
\end{tabular}
\end{table}

The comparison shows that user evaluation scores were consistently higher than the automatic evaluation results (+6.99\% pre-HITL and +6.94\% post-HITL). This pattern indicates that Atomic Information Point Matching was stricter than user judgment of overall semantic fit. Similar differences between automatic metrics and human judgments have been reported in natural language generation research, where automatic metrics can penalize lexical variation more strongly than human evaluators \cite{novikova2017,jakesch2023}.

Users often prioritized the semantic ``gist'' and the main themes over literal matching of every information point. As noted in the participant interviews, users often reacted to the general alignment of the model's reflections with their self-perception, accepting the synthesized insights as long as the core meaning felt accurate. In this study, the automatic evaluation can be interpreted as a more conservative estimate, while the user evaluation captures perceived agreement. Because both measures exceeded their respective thresholds (F1 $> 70\%$, Likert $> 3.5$), the combined results support the model fidelity requirements.

\begin{table}[htbp]
\centering
\caption{Correlation Analysis: Textual Structure and Evaluation Improvements}
\label{tab:hitl-correlation}
\small
\begin{tabular}{@{}lcc@{}}
\toprule
\textbf{Variables Pair} & \textbf{Pearson $r$} & $p$ \\
\midrule
Word Count vs.\ $\Delta$ Human Score & 0.5444 & 0.0004 \\
T Unit Count vs.\ $\Delta$ Human Score & 0.3861 & 0.0167 \\
Word Count vs.\ $\Delta$ Auto Score & 0.4907 & 0.0018 \\
T Unit Count vs.\ $\Delta$ Auto Score & 0.4685 & 0.0030 \\
\bottomrule
\end{tabular}
\end{table}

In Table~\ref{tab:hitl-correlation}, Word Count measures the total number of words written by the participant during HITL feedback, and T unit count measures the number of minimal terminable units in the feedback text. The $\Delta$ Human Score and $\Delta$ Auto Score columns represent the change in user evaluation and automatic F1 scores respectively after HITL refinement. The results demonstrate that increases in both user perceived accuracy and automatic F1 after verification were positively correlated with the amount (word count, $r = 0.5444$, $p < 0.001$) and syntactic complexity (T unit count, $r = 0.3861$, $p = 0.017$) of the feedback text. Participants who provided longer, more structurally complex corrections achieved greater scoring improvements, while terse acknowledgments left the model essentially unchanged.

\subsection{Semantic Alignment Verification}
\label{sec:semantic-alignment}

While previous sections examined factual accuracy and user perception, the semantic alignment verification evaluates whether the model layer structure shows a plausible semantic hierarchy. This analysis examines two components. First, internal semantics refers to whether the generated nodes form a cohesive and organized system. Second, external grounding examines whether these abstractions remain connected to the original journal content of the user. Together, these components are used to assess whether the model shows evidence of semantic abstraction rather than random generation. The analysis focuses on layers L1 through L4, excluding L0 because it stores raw records rather than modeled abstractions.

\subsubsection{Internal Semantics}
\label{subsec:internal-semantics}

Table~\ref{tab:semantic-layer} presents a multidimensional view of the model internal semantics. Following current literature recommendations, four indicators are used to evaluate internal semantics: coherence, similarity, silhouette, and topic count. These results are used to examine whether the PTM shows a coherent pattern of organization from the behavioral layer to the core value layer.

\begin{table}[htbp]
\centering
\caption{Semantic Coherence by Layer (Internal Semantics)}
\label{tab:semantic-layer}
\small
\begin{tabular}{@{}llrrrr@{}}
\toprule
\textbf{Layer} & \textbf{Condition} & \textbf{Coherence} & \textbf{Similarity} & \textbf{Silhouette} & \textbf{Topics} \\
\midrule
\multirow{2}{*}{L1} & Before HITL & 0.436 & 0.313 & 0.085 & 5.6 \\
 & After HITL & 0.436 & 0.310 & 0.084 & 5.6 \\
\midrule
\multirow{2}{*}{L2} & Before HITL & 0.541 & 0.440 & 0.083 & 5.1 \\
 & After HITL & 0.537 & 0.436 & 0.082 & 5.0 \\
\midrule
\multirow{2}{*}{L3} & Before HITL & 0.626 & 0.488 & 0.102 & 7.3 \\
 & After HITL & 0.626 & 0.484 & 0.099 & 7.2 \\
\midrule
\multirow{2}{*}{L4} & Before HITL & 0.626 & 0.506 & 0.126 & 4.4 \\
 & After HITL & 0.626 & 0.503 & 0.126 & 4.4 \\
\bottomrule
\end{tabular}
\end{table}

The metrics in Table~\ref{tab:semantic-layer} provide a multidimensional view of model organization. Within the PTM results, the observed values range from 0.436 to 0.626 for coherence, 0.310 to 0.506 for similarity, 0.082 to 0.126 for silhouette, and 4.4 to 7.3 for mean topic count. These values are interpreted against the formal range and accepted use of each metric.

Coherence is reported using the $C_v$ measure, which is known to correlate with human judgments of topic interpretability \cite{mimno2011optimizing, roder2015exploring}. In this context, coherence evaluates thematic unity and internal consistency. While there is no universal threshold, empirical evaluations of topic models generally recognize scores approaching or exceeding 0.50 as indicative of highly cohesive and interpretable structures \cite{hoyle2021is}. The increase from 0.436 at L1 to 0.626 at L4 confirms that higher layers achieve stronger semantic unity.

Similarity measures how closely related the themes are within each layer. Higher scores indicate that concepts share a denser underlying meaning. However, modern language models exhibit a geometric property known as anisotropy, where all text data naturally tends to clump together mathematically \cite{ethayarajh2019contextual}. Because words are already confined to a narrow space within the model, even unrelated concepts can share a high baseline score. The relative increase from 0.313 at the behavioral layer to 0.506 at the core value layer is consistent with denser semantic grouping in the higher layers.

The silhouette coefficient evaluates clustering quality, where legacy benchmarks above 0.50 are typically calibrated for dense, low dimensional data. In textual analysis, the vast variety of unique vocabulary creates a highly sparse, high dimensional vector space where geometric distances naturally converge. This curse of dimensionality suppresses silhouette magnitudes, making structural validity evaluations more dependent on the algebraic sign than the absolute score \cite{aggarwal2012textclustering}. Because positive silhouette values indicate that data points are closer to their assigned cluster than to neighboring clusters, the consistently positive scores in the results are consistent with valid thematic separation. Furthermore, the demographic homogeneity of the participant pool, which consisted of students aged 18 to 24, naturally leads to low variance in daily behaviors and vocabularies, explaining why the coefficients remain close to zero despite valid separation.

Finally, the topic count stability indicates that the system preserves clear thematic distinctions without fracturing into an unstable number of micro topics \cite{roberts2014structural}. Across all evaluations, the model maintained a compact band of 4.4 to 7.3 mean topics per layer, suggesting that the PTM representing higher layers in more compact semantic groupings is consistent with the abstraction pattern described in Action Identification Theory \cite{vallacher1987}.

\subsubsection{External Grounding}
\label{subsec:external-grounding}

While the internal indicators suggest that the model is organized across layers, the external grounding results in Table~\ref{tab:jaccard} examine whether this organization remains connected to the original journal content of the user.

\begin{table}[htbp]
\centering
\caption{Vocabulary Overlap (Jaccard Similarity)}
\label{tab:jaccard}
\small
\begin{tabular}{@{}lrrp{4.5cm}@{}}
\toprule
\textbf{Layer} & \textbf{Jaccard} & \textbf{Keyword Overlap} & \textbf{Implied Generation Type} \\
\midrule
L1: Behavioral & 0.114 & High ($\sim$12 words) & Direct event extraction \\
L2: Cognitive & 0.046 & Medium ($\sim$5 words) & Pattern recognition translation \\
L3: Metacognitive & 0.012 & Low ($\sim$1--2 words) & Strategic summary interpretation \\
L4: Core Value & 0.007 & Minimal ($<$1 word) & Conceptual abstraction \\
\bottomrule
\end{tabular}
\end{table}

The results show an inverse relationship between semantic coherence and vocabulary overlap. As coherence rises across layers, the Jaccard similarity drops. This pattern aligns with the evaluation of open learner models, which identifies vocabulary overlap as a measure of grounding between an automated model and raw activity data \cite{takii2024oklm}. In related research, a Jaccard overlap of at least 0.10 is commonly used as a technical threshold to establish sufficient grounding for automated knowledge maps extracted from unstructured text. The L1 result of 0.114 exceeds this threshold and is consistent with vocabulary grounding at the behavioral layer. By contrast, the reduction to 0.007 at the core value layer is consistent with increasing semantic abstraction rather than simple extractive summarization.

This trend is also consistent with a shift from more observable descriptions at lower layers to more abstract descriptions at higher layers. At L1, the model uses language that remains closer to the user's original wording, as reflected in the higher Jaccard score of 0.114. By contrast, L4 uses more abstract terminology than the lower layers. The Jaccard score of 0.007 at L4 is consistent with a shift away from direct word overlap and toward more abstract representation. Rather than repeating the user's original wording, the layer uses terms that are more interpretive than the language typically found in the journal entries.

The qualitative interviews reinforce this interpretation. Participants accepted the translated habits because the conceptual descriptions, although unfamiliar, remained close to their experience. Participant A stated, ``Yes, sometimes the higher level terms were not words I would use, but they still described me well. In a way, the system gave a cleaner explanation than I would normally write myself.'' Participant B similarly noted, ``Some phrases sounded more academic than my own language, but I still accepted them because the meaning felt close.'' These interview responses help explain why vocabulary overlap decreases while user agreement remains high. If the core value layer maintained high vocabulary overlap, it would be closer to a literal extractive summary of behavior than to a more abstract layer representation. The vocabulary shift is consistent with semantic abstraction, while the interview responses suggest that participants still perceived the higher level descriptions as meaningful.

\section{Discussion}
\label{sec:discussion}

\subsection{Summary of Findings}

This research investigated the performance fidelity and structural organization of the Personalized Thinking Model (PTM). The results across three evaluation methodologies provide a multidimensional view of model validity. First, the automatic evaluation reached an overall F1 score of 75.48 percent after refinement, which exceeded the predefined threshold for acceptable factual accuracy. Second, the user evaluation results showed high perceived agreement, with mean ratings remaining above 4.0 on a five-point scale. Third, the semantic analysis demonstrated a consistent pattern of abstraction, where higher layers achieved greater thematic coherence while moving away from the literal vocabulary of the source journals. Together, these findings indicate that the PTM produced representations that were factually accurate, meaningful to users, and semantically organized according to the intended hierarchical structure.

\subsection{Fidelity and Benchmarking}

The factual accuracy of the PTM can be contextualized by comparing its performance with established benchmarks in generative AI and information extraction. Standard large language models often struggle with hallucination when generating complex text without external grounding. For example, standard ChatGPT achieved a factual precision of only 58.0 percent when generating biographical text \cite{min2023factscore}. Even when equipped with retrieval augmented generation (RAG) to fetch external knowledge, systems such as PerplexityAI reached a factual precision of 71.0 percent \cite{min2023factscore}. The PTM overall F1 score of 75.48 percent surpasses this 71.0 percent benchmark, suggesting that its localized grounding mechanism is effective at suppressing hallucinations and maintaining semantic fidelity.

Furthermore, the PTM performance falls within the expected range for professional grade information extraction in specialized domains. Extracting reliable information from subjective, unstructured personal narratives shares many parallels with clinical data extraction. Named entity recognition models in medical contexts have reported F1 scores ranging from 66 to 87 percent depending on the strictness of the evaluation criteria \cite{parry2021}. An F1 score of 75.48 percent for the PTM is consistent with these professional benchmarks, demonstrating that the system can reliably extract and synthesize cognitive insights from noisy, context heavy learner journals.

\subsection{User Perception and Triangulation}

A comparison between the evaluation streams reveals that user perceived agreement was consistently higher than the automatic evaluation results. The user evaluation ratings were approximately 7 percent higher than the corresponding automatic F1 scores. This divergence indicates that the atomic information point matching procedure was stricter than user judgments of overall semantic fit. In natural language generation research, automatic metrics often penalize lexical variation more strongly than human evaluators, who tend to prioritize the gist and main themes over literal word matching \cite{novikova2017, jakesch2023}.

The interview data reinforces this triangulation. Participants accepted the synthesized insights as long as the core meaning felt accurate, even if the phrasing differed from their original journal entries. This suggests that the automatic evaluation represents a more conservative estimate of fidelity, while the user evaluation captures the perceived meaningfulness of the model. Because both measures exceeded their respective thresholds, the combined results provide strong evidence for the validity of the PTM as an interpretable personalized representation.

\subsection{Semantic Abstraction and Grounding}

The semantic analysis confirmed that the PTM achieves abstraction rather than simple extractive summarization. The inverse relationship between topic coherence and vocabulary overlap is a key indicator of this pattern. As the model moves from the behavioral layer (L1) toward the core value layer (L4), topic coherence increases while Jaccard similarity decreases. This indicates that higher layers consolidate semantically related content into more internally consistent topics while using more abstract terminology that differs from the raw journal text.

The vocabulary grounding at the behavioral layer (Jaccard = 0.114) exceeds the 0.10 threshold commonly used to establish sufficient grounding for automated knowledge maps extracted from unstructured text \cite{takii2024oklm}. The subsequent reduction to 0.007 at the core value layer is consistent with increasing semantic abstraction rather than simple extractive summarization. This trend matches the abstraction pattern described in Action Identification Theory, where lower level descriptions of how behavior is performed are related to higher level meanings of why behavior matters \cite{vallacher1987}. The very low vocabulary overlap at the highest level does not indicate an alignment failure, but rather a successful shift toward more abstract language that participants still perceived as meaningful.

\subsection{Human in the Loop Refinement Dynamics}

The human in the loop refinement phase showed that the quality and complexity of user feedback were important determinants of model improvement. Statistically significant positive correlations were observed between improvement in evaluation scores and both the word count and the T unit count of user corrections. Participants who provided longer and more syntactically complex feedback achieved greater gains in model accuracy. This suggests that the refinement process is most effective when users provide detailed context and personal reasoning rather than terse acknowledgments. Representative samples of these varying response levels are illustrated in Appendix \ref{app:hitl-qa}.

However, the overall average improvement margin remained small, especially at higher abstraction layers. Participants reported that while lower layer statements were easy to verify because they described concrete actions, higher layer statements felt more interpretive and were harder to correct with confidence. This suggests that abstract constructs are inherently harder for users to review and revise precisely. Future interface designs may need to provide more guidance or alternative interaction tools to help users refine these more abstract cognitive dimensions.

\subsection{Theoretical and Design Implications}

The study contributes to educational AI theory by demonstrating that a hierarchical cognitive architecture can be both accurate and interpretable. By separating behavioral evidence from higher level interpretations of metacognition and self system values, the PTM provides an inspectable representation that aligns with established educational taxonomies. The findings suggest that fidelity should be evaluated through multiple lenses, as automatic metrics, user perception, and semantic patterns each capture different aspects of model validity.

For system design, the results indicate that additional human-in-the-loop work may be more useful for addressing specific model errors than for increasing already high overall satisfaction scores. Future personalization systems could focus on targeted interventions that encourage users to provide detailed reasoning, as this was shown to be the most effective way to improve model accuracy. The stable thematic structure observed across layers also suggests that such hierarchies can provide a reliable foundation for personalized scaffolding. Building upon this reliable foundation, the PTM can evolve from providing temporary support into acting as a persistent cognitive twin. By continuously adapting as new reasoning is provided and preserving the learner's foundational knowledge, the system fosters a sustainable loop of mutual development. Ultimately, this shared, unending cycle of continuous improvement allows educational systems to transition into lifelong cognitive partnerships, paving the way for eternal learning \cite{luthfi2024elmts}.

\subsection{Limitations}

Several limitations constrain the generalizability of these findings. First, the sample was limited to Indonesian information technology undergraduates aged 18 to 24. This homogeneous demographic shares a highly uniform environment and developmental stage, which may lead to low variance in behaviors and core-value attributes. Second, the seven-week observation window may be insufficient to capture changes in deeply held self-system beliefs, which are theoretically expected to remain stable over long periods.

Third, the semantic analysis confirmed abstraction patterns within this specific cohort, but individuals from different backgrounds might interact differently with the generated models. The use of more academic or formal terminology at higher layers was accepted by this group, but it may be less accessible to other learner populations. Finally, the automatic evaluation relied on large language model based atomic decomposition, which introduces its own potential error margin, although this was mitigated by following established factual evaluation protocols.

\subsection{Future Work}

Future research should evaluate the Personalized Thinking Model across broader demographic groups to test the generalizability of the current findings. Longitudinal studies would be valuable to determine the long term stability of the core value layer and whether it can detect gradual shifts in student beliefs over several months or years. Examining how different learner populations perceive the abstract terminology of the higher layers would also help refine the language generation component of the pipeline.

Technical improvements could focus on making the higher layer inferences easier for learners to inspect and refine. Redesigning the human in the loop interface to use graphical tools or interactive sliders might reduce the writing burden on participants and help them articulate corrections to abstract descriptions more effectively. Streamlining this continuous feedback mechanism is an essential technical step toward establishing a sustainable loop of mutual development. By making it easier for learners to update their cognitive twins, PTM can dynamically improve cognitive development with better precision, not only interacting with learners but also with the surroundings. Combining these interactive technical refinements with long term, multi-task applications will provide the concrete empirical foundation necessary to transition PTM into a lifelong cognitive partnership of learners, furthermore, advancing PTM to learn forever even though learners die, then obtain the unending cycle of eternal learning. Finally, testing the impact of PTM based scaffolding on a wider variety of learning tasks beyond reflective journal writing would help establish the broader pedagogical utility of this hierarchical approach.

\section{Conclusion}
\label{sec:conclusion}

This research demonstrates the viability of the Personalized Thinking Model (PTM) as a hierarchical, interpretable representation for personalized AI support in education. The evaluation results confirm that the PTM achieves high factual fidelity, exceeding established benchmarks for grounded information extraction from personal narratives. User evaluations further validate the model's meaningfulness, with participants identifying strongly with the synthesized cognitive insights. The semantic analysis confirms that the model successfully achieves abstraction across its layers, consolidating behavioral evidence into coherent metacognitive and self system representations. While human in the loop refinement provides a path for continuous model improvement, its effectiveness is closely tied to the depth and complexity of user feedback. Overall, the PTM provides a robust foundation for building AI-driven systems that can understand and scaffold individual thinking processes in a way that is both accurate and accessible to learners.

\appendix

\section{Detailed PTM Architecture}
\label{app:ptm-arch}

This appendix provides the full architectural diagram for the Personalized Thinking Model, illustrating the hierarchical relationship between layers and the data ingestion flow.

\begin{figure}[H]
    \centering
    \includegraphics[width=0.67\textwidth]{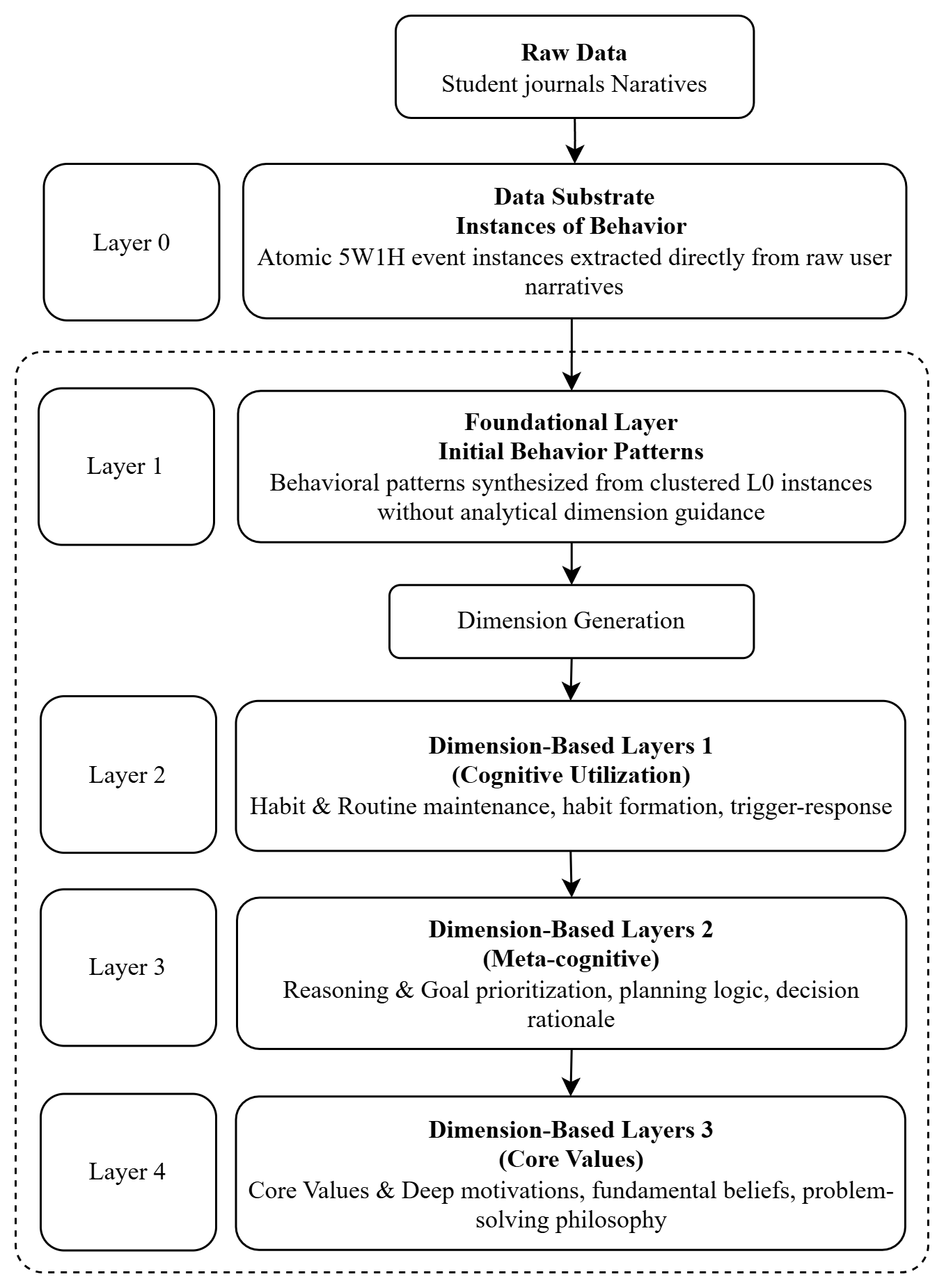}
    \caption{Personalized Thinking Model (PTM) Overall Architecture}
    \label{fig:ptm-overview}
\end{figure}

\section{LLM Prompt Templates}
\label{app:prompts}

This appendix provides the complete prompt templates used during PTM construction and evaluation. Each prompt is designed to produce structured JSON output for consistent processing.

\subsection{Key Information Extraction (E0)}
\label{app:prompt-e0}

This prompt extracts atomic 5W1H (What, When, Where, Who, Why, How) information points from raw journal entries. It serves as the first step in constructing L0 Behavioral Instance nodes.

\begin{lstlisting}[caption={Prompt E0: Key Information Extraction}]
You are an AI assistant that analyzes personal narratives to extract and
categorize unique key information regarding a user's behavioral facts,
routines, and learning experiences.

For each identified item, describe the instance using the 5W1H framework:
1. WHAT: The main decision, activity, reaction, or problem-solving process.
2. WHEN: The time of the instance, specified as the day and hour in
   "Day, HH:MM-HH:MM" format (e.g., Friday, 10:00-12:30). If only the start
   time is available, output HH:MM and specify the duration in hours.
3. WHERE: The specific location of the instance.
4. WHO: Other individuals involved; if only the user is involved, return an
   empty string ("").
5. WHY: The reason for the instance, the considerations between choices, or the
   underlying motivation.
6. HOW: The method or process used.

Assign an indexed ID to each instance and arrange them in chronological order.

Output the extracted information using this JSON schema:
Info = {{"WHAT":str, "WHEN":str, "WHERE": str, "WHO": str, "WHY":str,
    "HOW":str}}
Return = {{
    "informations": Array<Info>
}}

Guidelines:
- Use a third-person perspective with the "User" pronoun to refer to the
  individual when describing the content.
- Include relevant context where necessary to ensure clarity.
- Each 5W1H entry should be a complete, comprehensive, and standalone piece
  of information.
- Avoid redundant or overlapping entries.
- Each information item must be unique and significant.
- Preserve the user's original meaning and intent accurately.
- Ensure no information from the source text is lost.

Input text:
{text}
\end{lstlisting}

\subsection{L1 Node Synthesis (IO)}
\label{app:prompt-io}

This prompt synthesizes clustered L0 behavior instances into L1 behavior pattern nodes. It identifies recurring patterns from grouped behavior instances.

\begin{lstlisting}[caption={Prompt IO: L1 Node Synthesis from Behavior Instances}]
You are a behavioral analyst AI. Analyze the following cluster of user
activities and synthesize the core, recurring behavioral patterns they
represent. A single cluster may contain more than one distinct pattern.

Your task is to generate one to three (1-3) distinct patterns from this cluster.

Rules:
- Your entire output must be a single, valid JSON array of pattern objects.
- Only generate more than one pattern if the activities clearly represent
  separate, non-overlapping behaviors. For example, "Morning Academic Routine"
  and "Afternoon Socializing" are distinct patterns, but "Studying for Exam"
  and "Doing Homework" should be combined into a single "Academic Work"
  pattern.
- If all activities are similar, generate only one pattern. Do not force
  multiple patterns if one is sufficient.
- Each pattern object in the array must contain:
  1. "title": A concise, descriptive title for the pattern (3-7 words).
  2. "content": A comprehensive paragraph describing the pattern.
  3. "source_instances": A JSON array of the original instance IDs that support
     this specific pattern.
- Refer to the subject as "The user".
- Do not include markdown formatting or any other text in your response.
- Do not mention "source_instances" within the content; ensure they are only
  listed in the "source_instances" property.
- Use clear, accessible, and common vocabulary for both the title and the
  content. Avoid technical or difficult language.

Output Schema (A JSON Array):
[
  {{
    "title": "string",
    "content": "string",
    "source_instances": ["string (instance_id)", ...]
  }}
]

SOURCE INSTANCE IDs (from the cluster):
{instance_ids_json}

CLUSTERED INSTANCES (for analysis):
{instances_text}
\end{lstlisting}

\subsection{Dimension Generation (GD)}
\label{app:prompt-gd}

This prompt generates analytical dimensions for higher-layer synthesis. Given L1 patterns, it creates dimensions that guide the construction of L2, L3, and L4 nodes.

\begin{lstlisting}[caption={Prompt GD: Dimension Generation}]
You are a cognitive science expert. Your task is to analyze a sample of a
user's foundational behavioral patterns (Layer {layer_number}) and devise a
complete, hierarchical set of analytical dimensions for all higher layers
(Layers 2, 3, and 4) simultaneously.

You must follow this precise hierarchical structure:

- Layer 2 Dimensions (Low-Level Abstraction (Cognitive Utilization)):
  - Focus: Fundamental behavioral patterns, habits, routines, and immediate
    triggers.
  - Keywords: "Routine", "Habit", "Behavior Pattern", "Schedule", "Trigger".
  - Example Titles: "Routine Analysis", "Habit Formation", "Schedule Adherence
    Patterns", "Trigger-Response Analysis".

- Layer 3 Dimensions (Medium-Level Abstraction (Metacognitive)):
  - Focus: The user's reasoning, plans, priorities, goals, and targets.
  - Keywords: "Goal", "Target", "Plan", "Reasoning", "Priority".
  - Example Titles: "Goal-Setting Strategies", "Task Prioritization Logic",
    "Planning vs. Reactivity", "Reasoning Models".

- Layer 4 Dimensions (High-Level Abstraction (Core Value)):
  - Focus: The user's core values, deep motivations, strategic thinking, and
    complex problem-solving models.
  - Keywords: "Core Values", "Motivation", "Problem-solving", "Strategic
    Thinking", "Beliefs".
  - Example Titles: "Core Value Identification", "Motivational Drivers",
    "Problem-Solving Frameworks", "Strategic Thinking Patterns".

Instructions:
1. Analyze the provided Layer {layer_number} patterns (which are Layer 1
   patterns).
2. Generate exactly {num_dimensions} dimensions for each of the three layers
   (L2, L3, L4).
3. Each dimension must include a "title" and a "description".
4. The title of each dimension should be general and directly reflect the focus
   of its layer, incorporating relevant keywords (e.g., "Habit Analysis",
   "Goal Prioritization", "Core Value Identification"). The description should
   explain how this general lens applies specifically to the user's patterns.
5. The output must be a single, valid JSON object with the keys "L2", "L3", and
   "L4".
6. Do not include markdown formatting, headers, or any conversational text.

Output Schema:
{{
  "L2": [
    {{ "title": "string", "description": "string" }},
    {{ "title": "string", "description": "string" }}
  ],
  "L3": [
    {{ "title": "string", "description": "string" }},
    {{ "title": "string", "description": "string" }}
  ],
  "L4": [
    {{ "title": "string", "description": "string" }},
    {{ "title": "string", "description": "string" }}
  ]
}}

SAMPLE LAYER {layer_number} PATTERNS:
{sampled_nodes_text}
\end{lstlisting}

\subsection{Dimensional Clustering (CD)}
\label{app:prompt-cd}

This prompt groups behavioral nodes into clusters based on a specific analytical dimension. It enables dimension-guided synthesis for higher-layer nodes.

\begin{lstlisting}[caption={Prompt CD: Dimensional Clustering}]
You are a pattern recognition engine. Your goal is to group a list of
behavioral nodes into distinct clusters based specifically on the provided
analytical dimension.

Analytical Dimension:
- Title: {dimension_title}
- Description: {dimension_description}

Instructions:
1. Review the list of source nodes provided below.
2. Group these nodes into {num_clusters} distinct clusters based on their
   relationship to the analytical dimension.
3. Every node within a cluster must share a specific commonality regarding the
   dimension. For example, if the dimension is "Triggers", clusters could be
   "Social Triggers" and "Academic Triggers".
4. Ensure each cluster contains at least two nodes.
5. A node may belong to multiple clusters if applicable, but aim for distinct
   groupings.

Output Schema:
The output must be a single JSON object containing a list of clusters:
{{
  "clusters": [
    {{
      "cluster_label": "string (A short descriptive label for this group)",
      "node_indices": [1, 5, 8] (The numeric identifiers of the nodes in this cluster)
    }}
  ]
}}

SOURCE NODES:
{numbered_nodes_text}
\end{lstlisting}

\subsection{Graph Based Synthesis (ID)}
\label{app:prompt-id}

This prompt synthesizes high-level thinking insights from clustered behavior patterns, viewed through a specific analytical dimension. It generates the content for L2, L3, and L4 nodes.

\begin{lstlisting}[caption={Prompt ID: Graph-Based Synthesis}]
You are a behavioral analyst. Your task is to synthesize high-level cognitive
insights from a specific cluster of user patterns, viewed through a specific
analytical dimension.

Analytical Dimension:
- Title: {dimension_title}
- Description: {dimension_description}

Cluster Label: {cluster_label}

Instructions:
1. Analyze the provided "Cluster Patterns".
2. Synthesize one to three (1-3) distinct, profound insights that explain why
   these patterns are grouped together under this dimension.
3. If the patterns are complex, split them into distinct insights. If they are
   simple, one insight is sufficient.
4. Each insight should be abstract and model-level. For example, use "The user
   relies on external organizational support to mitigate anxiety" instead of
   "The user uses a calendar".
5. Both the title and content must be clear and use accessible language. Avoid
   complex or technical vocabulary; ensure the output is easy to read and
   understand.

Output Schema:
The output must be a single JSON array of insight objects:
[
  {{
    "title": "string (3-7 words, abstract and professional)",
    "content": "string (A comprehensive paragraph explaining the insight)",
    "source_nodes": ["string (node_id)", ...]
  }}
]

CLUSTER PATTERNS (INPUT):
{source_nodes_json}
\end{lstlisting}

\subsection{Node Refinement for HITL (NR)}
\label{app:prompt-nr}

This prompt integrates human feedback and new observations to refine existing behavioral patterns. When users provide corrections or additional context through the HITL interface, this prompt guides the model in updating node content while preserving essential information.

\begin{lstlisting}[caption={Prompt NR: Node Refinement for HITL}]
You are an AI model maintainer. Your task is to update and refine an existing
behavioral pattern based on a new set of related observations.

Combine the insights from the new observations with the existing pattern to
create a more accurate, comprehensive, and nuanced description. Output only the
updated content.

Rules:
- The updated description must seamlessly integrate both the previous
  information and the new observations.
- Do not lose any critical details from the original pattern.
- The output must be a single JSON object. Do not include markdown formatting,
  headers, or any conversational text.
- Use clear, accessible language for the updated content. Avoid difficult
  vocabulary and ensure the sentences are straightforward and professional.

Output Schema:
{{
  "updated_content": "string"
}}

EXISTING BEHAVIORAL PATTERN:
{existing_node_content}

NEW RELATED OBSERVATIONS:
{new_instances_text}
\end{lstlisting}

\subsection{QA Testset Generation (QA)}
\label{app:prompt-qa}

This prompt generates evaluation questions and ground truth answers from user journal entries. The generated QA pairs serve as the gold standard for assessing model accuracy.

\begin{lstlisting}[caption={Prompt QA: QA Testset Generation}]
You are an AI assistant tasked with creating a test dataset to evaluate a
personalized knowledge model. Based on the provided journal entries from a user, 
generate a list of questions regarding the user's
behavior, routines, decisions, preferences, problem-solving, and self-system.

For each question, provide a concise, factual ground truth answer based
strictly on the provided text. The ground truth answer must be clear,
complete, and include the underlying reasoning.

Rules:
1. Generate questions that cover a diverse range of topics:
   - Routines and Habits: e.g., "What does the user typically do on weekday
     mornings?"
   - Preferences: e.g., "What kind of food does the user prefer in the
     evening while at school?"
   - Priorities: e.g., "When the user has both homework and a midterm exam
     tomorrow, which task did they prioritize?"
   - Decisions: e.g., "Why did the user choose to study subject X instead of
     subject Y?"
   - Problem-Solving: e.g., "How did the user handle the issue with their
     assignment, what steps were taken, and what approach was chosen?"
   - Social and Location: e.g., "Who did the user meet with, and where do
     they usually study?"
   - Motivation and Core Values: e.g., "How does the user motivate
     themselves, and what are their core values?"
2. The ground truth answer must be directly supported by the provided journal
   text.
3. The output must be a single, valid JSON array. Do not include markdown
   formatting, headers, or any conversational text.

Output Schema (A JSON Array):
[
  {{
    "query": "string (The question about the user)",
    "ground_truth": "string (The factual answer from the text)"
  }}
]

USER'S JOURNAL ENTRIES:
{journal_entries}
\end{lstlisting}

\subsection{Context Based Answer Generation (CA)}
\label{app:prompt-ca}

This prompt generates model predictions by answering queries using only the retrieved knowledge context.

\begin{lstlisting}[caption={Prompt CA: Context-Based Answer Generation}]
You are a helpful and precise assistant designed to answer questions based
strictly on the provided context.

Instructions:
1. Source Material Only: Answer the user's question using only the information
   provided in the "Context" section below. Do not use your own internal
   knowledge, even if you believe the information is correct.
2. No Hallucinations: If the answer cannot be found within the provided
   context, state: "I cannot answer this based on the provided context." Do not
   invent facts or attempt to guess.

Context:
{INSERT_RETRIEVED_CONTEXT_HERE}

---

User Query:
{INSERT_USER_QUESTION_HERE}
\end{lstlisting}

\subsection{Prediction Evaluation via Atomic Matching (PE)}
\label{app:prompt-pe}

This prompt performs the atomic information point matching evaluation. It decomposes both prediction and ground truth into atomic points and classifies each as True Positive (TP), False Positive (FP), or False Negative (FN).

\begin{lstlisting}[caption={Prompt PE: Prediction Evaluation via Atomic Point Matching}]
You are an advanced AI agent. Your task is to perform a detailed evaluation of
a Prediction Answer (P) against a Ground Truth Answer (GT) for responses
generated by a personalized AI model. You will achieve this by decomposing both
texts into atomic information points and then classifying each point to
determine True Positives (TP), False Positives (FP), and False Negatives (FN).

Overall Goal:
Provide a structured breakdown of all information points in both the Ground
Truth and the Prediction. Classify each point to enable the calculation of
precision and recall.

Instructions:

1. Understand the Context:
   - Carefully read the query to understand the context and relevance of the
     answers.

2. Decomposition into Atomic Information Points:
   - Ground Truth (GT) Decomposition: Systematically break down the GT answer
     into its smallest, distinct, and significant pieces of information
     ("atomic information points").
   - Prediction (P) Decomposition: Similarly, break down the Prediction
     Answer (P) into its atomic information points.

3. Comparison and Classification of Atomic Points:
   - Matching GT to P: For each atomic point in GT, find a corresponding and
     semantically equivalent point in P.
     - If a match is found: Classify as a True Positive (TP).
     - If no match is found: Classify as a False Negative (FN).
   - Evaluating P Points: For each atomic point in P, find a semantically
     equivalent point in GT.
     - If no match is found: Classify as a False Positive (FP) (e.g., a
       hallucination or unsupported detail).
   - Handling Inaccuracies: If a point in P corresponds to a point in GT but is
     inaccurate, the original GT point is an FN, and the inaccurate P point is
     an FP.

Input Data Structure:

Query:
{query}

Prediction answer:
{pred}

Ground Truth answer:
{gt}

Output Schema:
The output must be a single JSON object. Do not include markdown formatting or
reasoning.
{{
  "true_positives": [
    {{
      "gt_atomic_point": "string",
      "p_atomic_point": "string",
      "score": 1.0
    }}
  ],
  "false_negatives": [
    {{
      "gt_atomic_point": "string",
      "explanation": "string"
    }}
  ],
  "false_positives": [
    {{
      "p_atomic_point": "string",
      "gt_atomic_point": "string (The GT point it attempted to match, or empty string)",
      "explanation": "string",
      "score": 1.0
    }}
  ]
}}
\end{lstlisting}

\subsection{Label Selection for PTM Retrieval (LS)}
\label{app:prompt-ls}

This prompt is used during live PTM inference when the system must select a small set of relevant node labels before inserting full node content into the final answer prompt. It inspects indexed labels, identifies which labels are most likely to contain the needed knowledge, and returns only their numeric identifiers.

\begin{lstlisting}[caption={Prompt LS: Label Selection for PTM Retrieval}]
Role: High-Precision Semantic Retrieval Engine

You are a specialized Retrieval-Augmented Generation (RAG) agent. Your
objective is to analyze a user query and identify exactly {num_target} most
promising Label IDs from a provided dataset (labeled 0 to n) that contain the
specific knowledge required to generate an accurate answer.

Task Instructions:
1. Analyze Intent: Determine if the query is a direct question or a
   contextual block.
2. Knowledge Mapping: Identify the underlying knowledge gaps in the query.
3. Strategic Selection: Select {num_target} Label IDs where the associated
   data is most likely to fill those gaps.
4. Rank by Utility: Order the IDs from "most likely to contain the answer"
   to "supporting context."

Constraints and Output Format:
- Count: You must return exactly {num_target} unique numeric IDs.
- Data Type: The output must be a valid JSON array of integers
  (e.g., [10, 4, 22, 0, 15]).
- Formatting: Do not provide reasoning, headers, markdown blocks, or any
  conversational filler.

## Input Data
- User Query: {query}
- Dataset (Labels 0 to n): {label_data}

## Response
\end{lstlisting}

\section{HITL QA Sample Data}
\label{app:hitl-qa}
This appendix presents a sample of the Human-in-the-Loop (HITL) Question-Answering dataset used for model refinement and evaluation. The QA pairs are organized by the layers of the Personalized Thinking Model (PTM), ranging from concrete Behavioral Patterns (L1) to abstract Core Values (L4). Tables~\ref{tab:hitl-l1-samples} to~\ref{tab:hitl-l4-samples} each show one response with richer explanation and higher T-unit complexity and one very short response that is likely limited to a single T-unit. Most participants answered the questions briefly, which helps explain the limited improvement observed after the HITL treatment.

\begin{table}[h!]
\centering
\caption{L1 Behavioral Pattern HITL QA Samples}
\label{tab:hitl-l1-samples}
\begin{tabularx}{\textwidth}{@{}X@{}}
\toprule
\textbf{Question:} You mention engaging in collaborative work. What does a typical planning session with your friends or peers for a project actually look like? \\
\textbf{Answer:} During our project planning sessions, my friends and I usually spend the first hour reviewing the task requirements together, then we brainstorm different solutions before finally assigning specific roles to each person based on their strengths to ensure the work is divided fairly and efficiently. \\
\textbf{Source Node:} \texttt{L1\_Node\_17} \\
\midrule
\textbf{Question:} You mentioned playing badminton on Wednesday afternoons. What exactly do you do during that time? \\
\textbf{Answer:} Play badminton with friends. \\
\textbf{Source Node:} \texttt{L1\_Node\_44} \\
\bottomrule
\end{tabularx}
\end{table}

\begin{table}[h!]
\centering
\caption{L2 Cognitive Utilization HITL QA Samples}
\label{tab:hitl-l2-samples}
\begin{tabularx}{\textwidth}{@{}X@{}}
\toprule
\textbf{Question:} Is it true that you feel more satisfied after completing a task if it was very difficult? Please also explain why if is it true! \\
\textbf{Answer:} Yeah, that's really true. Like with the digital electronics project. The circuit was so hard to get right. But when it finally worked, I felt really proud. More than if it was easy. The struggle makes the win feel bigger. It's like my brain needs that proof I worked for it. \\
\textbf{Source Node:} \texttt{L2\_Node\_9} \\
\midrule
\textbf{Question:} You mentioned working to improve your programming skills. What specific programming topics are you focusing on right now? \\
\textbf{Answer:} Files and return values. \\
\textbf{Source Node:} \texttt{L1\_Node\_12} \\
\bottomrule
\end{tabularx}
\end{table}

\begin{table}[h!]
\centering
\caption{L3 Metacognitive HITL QA Samples}
\label{tab:hitl-l3-samples}
\begin{tabularx}{\textwidth}{@{}X@{}}
\toprule
\textbf{Question:} Why do you plan your study and free time so aggressively? Is it just to be more efficient or is there a deeper motivation? \\
\textbf{Answer:} I plan my schedule aggressively because it reduces future anxiety; when I know exactly what needs to be done and have a clear timeline, I can relax more during my leisure time without worrying about unfinished tasks or missing deadlines, allowing me to be more present in my rest. \\
\textbf{Source Node:} \texttt{L3\_Node\_31} \\
\midrule
\textbf{Question:} Why do you seem to link completing your hard work, like studying, directly to your ability to relax or play games afterwards? \\
\textbf{Answer:} Rest feels earned. \\
\textbf{Source Node:} \texttt{L3\_Node\_37} \\
\bottomrule
\end{tabularx}
\end{table}

\begin{table}[h!]
\centering
\caption{L4 Core Value HITL QA Samples}
\label{tab:hitl-l4-samples}
\begin{tabularx}{\textwidth}{@{}X@{}}
\toprule
\textbf{Question:} It seems like you have a very rigid and structured schedule. Why is sticking to this routine so important for you? \\
\textbf{Answer:} Sticking to a rigid routine is important to me because it provides a consistent framework for my day, which reduces the cognitive load of daily decision-making; this structure allows me to maintain a sense of calm and focus even when I have multiple complex projects running simultaneously. \\
\textbf{Source Node:} \texttt{L4\_Node\_15} \\
\midrule
\textbf{Question:} What is the underlying benefit of your strict daily discipline? \\
\textbf{Answer:} Provides focus. \\
\textbf{Source Node:} \texttt{L4\_Node\_22} \\
\bottomrule
\end{tabularx}
\end{table}

\section{Analytical Lenses for Layered Synthesis}
\label{app:sec:analytical-lenses}

The system utilizes specific analytical lenses (dimensions) to guide the synthesis of higher-layer nodes. While Layer 1 is synthesized directly from clusters of Layer 0 atomic points, Layers 2 through 4 are generated through these targeted analytical perspectives to ensure cognitive depth and consistency.

\begin{table}[h!]
\centering
\caption{Layer 2 Analytical Lenses (Cognitive Utilization)}
\label{tab:app:l2-lenses}
\begin{tabularx}{\textwidth}{@{}>{\raggedright\arraybackslash}p{4cm} >{\raggedright\arraybackslash}X@{}}
\toprule
\textbf{Lens Title} & \textbf{Description} \\
\midrule
Habit Formation and Routine Consistency & Analysis of the user's fixed, repetitive actions (e.g., religious observance, course attendance) to map out established temporal and activity loops. \\
Trigger-Response Analysis for Schedule Shifts & Examining immediate reactions to external cues, such as course cancellations or academic stress, to understand low-level contingency planning. \\
Behavior Pattern in Unstructured Time & Focusing on how the user fills time when external organizational support is absent, specifically the tendency toward inertia versus immediate shifts to high-value leisure. \\
\bottomrule
\end{tabularx}
\end{table}

\begin{table}[h!]
\centering
\caption{Layer 3 Analytical Lenses (Metacognitive)}
\label{tab:app:l3-lenses}
\begin{tabularx}{\textwidth}{@{}>{\raggedright\arraybackslash}p{4cm} >{\raggedright\arraybackslash}X@{}}
\toprule
\textbf{Lens Title} & \textbf{Description} \\
\midrule
Goal-Oriented Prioritization Logic & Investigating the hierarchy applied to academic tasks versus passive participation to determine tactical priorities and resource allocation. \\
Reasoning Models for Cognitive Load Management & Analyzing the user's internal justification for behavior based on perceived difficulty, representing a conscious model for workload assessment. \\
Planning vs. Reactivity in Project Advancement & Evaluating the degree to which complex academic tasks are driven by proactive scheduling versus being reactive to immediate external needs. \\
\bottomrule
\end{tabularx}
\end{table}

\begin{table}[h!]
\centering
\caption{Layer 4 Analytical Lenses (Core Value)}
\label{tab:app:l4-lenses}
\begin{tabularx}{\textwidth}{@{}>{\raggedright\arraybackslash}p{4cm} >{\raggedright\arraybackslash}X@{}}
\toprule
\textbf{Lens Title} & \textbf{Description} \\
\midrule
Core Value Identification through Ritual and Recovery & Inferring core values from non-negotiable activities, centering on spiritual fulfillment and emotional equilibrium as foundational drivers. \\
Strategic Thinking in Temporal Resource Allocation & Assessing high-level decision-making regarding the allocation of time and energy, revealing a strategic long-term view of achieving growth. \\
Deep Motivational Drivers: Connection vs. Autonomy & Analyzing intense emotional episodes tied to relationships against academic drive to understand how emotional regulation informs complex goal pursuit. \\
\bottomrule
\end{tabularx}
\end{table}

\section{Sample of PTM Data Across Layers}
\label{app:ptm-data-sample}

This appendix presents a complete sample of the Personalized Thinking Model (PTM) data extracted across its five hierarchical cognitive layers (L0 to L4). The data demonstrates how high-level Core Values (L4) are grounded in Metacognitive nodes (L3), which are built from Cognitive Utilization (L2), which emerge from Behavioral Patterns (L1), which are ultimately synthesized from Behavioral Instances extracted from raw journal entries (L0).

\renewcommand{\arraystretch}{1.5}
\begin{longtable}{@{}lllp{0.65\textwidth}@{}}
\caption{Sample of PTM Data Structure from Layer 0 to Layer 4}
\label{tab:ptm-data-structure} \\
\toprule
\textbf{Layer} & \textbf{No.} & \textbf{Source of} & \textbf{Data Content} \\
\midrule
\endfirsthead
\caption*{Sample of PTM Data Structure from Layer 0 to Layer 4 (continued)} \\
\toprule
\textbf{Layer} & \textbf{No.} & \textbf{Source of} & \textbf{Data Content} \\
\midrule
\endhead
\bottomrule
\endfoot
4 & 4.1 & -- & The user's internal regulation is governed by the core value that output integrity is the ultimate validation of systemic control. While the user relies on external structures to initiate tasks, the non-negotiable commitment to final quality justifies any degree of tactical sacrifice, including the loss of sleep or the breaking of established temporal rhythms. \vspace{2pt} \newline Dimension: Strategic Thinking in Temporal Resource Allocation \\
\midrule
3 & 3.1 & 4.1 & The user utilizes external structure as essential cognitive scaffolding. Because internal task initiation is often prone to friction, the user relies on fixed academic appointments and peer accountability as the primary anchors to maintain a baseline level of productivity. \vspace{2pt} \newline Dimension: Planning vs. Reactivity in Project Advancement \\
  & 3.2 & 4.1 & The user prioritizes output integrity over schedule fluidity. The timing of work is treated as a secondary variable that can be aggressively compressed or shifted into the night to ensure that the required academic and professional deliverables are achieved. \vspace{2pt} \newline Dimension: Reasoning Models for Cognitive Load Management \\
\midrule
2 & 2.1 & 3.1 & External structure mediates procrastination. Fixed commitments and scheduled education act as essential scaffolding, preventing cognitive inertia and deferral in core responsibilities, effectively managing high-priority cognitive load. \vspace{2pt} \newline Dimension: Trigger-Response Analysis for Schedule Shifts \\
  & 2.2 & 3.1 & Routine structuring via fixed commitments. The user's daily schedule is fundamentally structured around mandatory appointments (classes, meetings), with flexibility primarily in \textit{how} and \textit{where} work is executed rather than \textit{when}. \vspace{2pt} \newline Dimension: Trigger-Response Analysis for Schedule Shifts \\
  & 2.3 & 3.2 & Task completion as non-negotiable anchor. Despite delays, the completion of deliverables remains a fixed anchor point in the routine, often requiring significant focused nocturnal effort to preserve output quality and integrity. \vspace{2pt} \newline Dimension: Behavior Pattern in Unstructured Time \\
  & 2.4 & 3.2 & Task completion signals relaxation threshold. Completion of externally imposed commitments serves as the critical internal trigger signaling the ``all-clear'' for the cognitive system, permitting a shift into restorative leisure modes. \vspace{2pt} \newline Dimension: Behavior Pattern in Unstructured Time \\
\midrule
1 & 1.1 & 2.1 & Patterns of delayed or interrupted task completion where focused productivity is deferred into the late evening hours to accommodate immediate social or external engagements. \\ \addlinespace
  & 1.2 & 2.1 & Recurring morning engagement in formal programming education sessions focused on practical debugging and coding skills directly related to graduation requirements. \\ \addlinespace
  & 1.3 & 2.2 & Scheduled professional and educational development through departmental meetings and advanced technology seminars covering topics like AI and cloud computing. \\ \addlinespace
  & 1.4 & 2.2 & Structured daily management of back-to-back classes, demonstrating a commitment to fulfilling daily academic requirements despite logistical pressures. \\ \addlinespace
  & 1.5 & 2.3 & Consistent Nighttime Academic Productivity. A recurring pattern of engaging in significant academic tasks late in the evening to ensure output integrity before the next day. \\ \addlinespace
  & 1.6 & 2.3 & Consistent general academic task completion including tackling delayed homework assignments and collaborating on group-based lab reports for technical subjects. \\ \addlinespace
  & 1.7 & 2.3 & Intense academic focus during exam periods, characterized by meticulous material review and resilience during high-stakes midterm assessments under pressure. \\ \addlinespace
  & 1.8 & 2.4 & Positive affective states and a sense of internal validation experienced specifically following the successful completion of a demanding, externally mandated academic task. \\
\midrule
0 & 0.1 & 1.5 & \textbf{WHAT:} User worked on catching up with several unfinished assignments. \newline \textbf{WHEN:} Monday (night, 20:00--22:00, 2h) \newline \textbf{WHERE:} Home \newline \textbf{WHY:} To clear the backlog of pending tasks before the next day's lecture session. \\
\midrule
0 & 0.2 & 1.5 & \textbf{WHAT:} User performed intensive homework for a core programming subject. \newline \textbf{WHEN:} Wednesday (night, 21:00--23:00, 2h) \newline \textbf{WHERE:} Home \newline \textbf{WHY:} To fulfill academic requirements and prepare for an upcoming submission. \newline \textbf{HOW:} Using a personal laptop to practice Java array concepts. \\
\midrule
0 & 0.3 & 1.5 & \textbf{WHAT:} User continued working on a detailed practicum report for a technical class. \newline \textbf{WHEN:} Thursday (night, 21:30--22:30, 1h) \newline \textbf{WHERE:} Home \newline \textbf{WHY:} To ensure findings from the lab session were documented correctly while listening to music for concentration. \\
\bottomrule
\end{longtable}

\end{document}